\documentclass[lettersize,journal]{IEEEtran}

\IEEEoverridecommandlockouts                              %

\usepackage{etoolbox}
\usepackage{capt-of}
\usepackage{url}
\makeatletter
\let\NAT@parse\undefined
\makeatother
\usepackage{graphics} %
\usepackage{epsfig} %
\usepackage{times} %
\usepackage{mathtools,amsmath} %
\usepackage{amssymb}  %
\usepackage[bookmarks=true]{hyperref}
\usepackage{xcolor,colortbl}
\usepackage{color}
\usepackage{siunitx}
\usepackage{multirow}
\usepackage{multicol}
\usepackage[skip=2pt,font=small]{caption}
\usepackage{subcaption}
\usepackage{booktabs}
\usepackage{pifont}
\usepackage{cite}
\usepackage{bm}
\usepackage[ruled,vlined]{algorithm2e}
\usepackage{tabularx}
\usepackage{balance}
\usepackage{cancel}
\usepackage{multirow, caption}

\usepackage{tikz}
\usetikzlibrary{svg.path,external}
\usepackage[utf8]{inputenc}
\usepackage{pgfplots}
\DeclareUnicodeCharacter{2212}{−}
\usepgfplotslibrary{groupplots,dateplot}
\usetikzlibrary{patterns,shapes.arrows}
\pgfplotsset{compat=newest}

\newcommand{\extension}[1]{{#1}}
\newcommand{\rev}[1]{\begingroup#1\endgroup}
\newcommand{\revi}[1]{\begingroup#1\endgroup}

\newcommand{\final}[1]{{\color{black}#1}}

\newcommand{\hide}[1]{}
\let\vec\bm
\newcommand{\mat}[1]{\begin{bmatrix}#1\end{bmatrix}}

\DeclareMathOperator*{\argmin}{argmin}
\let\vec\bm

\definecolor{Gray}{gray}{0.9}
\definecolor{somegray}{rgb}{0.5, 0.5, 0.5}
\newcommand{\darkgrayed}[1]{\textcolor{somegray}{#1}}

\makeatletter
\newcommand*\titleheader[1]{\gdef\@titleheader{#1}}
\AtBeginDocument{%
  \let\st@red@title\@title
  \def\@title{%
    \vskip-2em
    \bgroup\normalfont\large\centering\@titleheader\par\egroup
    \vskip1.5em\st@red@title}
}
\makeatother

\titleheader{ \darkgrayed{This paper has been accepted for publication at \\ IEEE Transactions on Robotics 2025 
\copyright IEEE} }

\title{\LARGE \bf Actor-Critic Model Predictive Control: Differentiable Optimization meets Reinforcement Learning \revi{for Agile Flight}}

\author{Angel Romero, Elie Aljalbout, Yunlong Song, Davide Scaramuzza
    \thanks{
    The authors are with the Robotics and Perception Group, University of Zurich, Switzerland (\protect\url{http://rpg.ifi.uzh.ch}). This work was supported by the European Union’s Horizon Europe Research and Innovation Programme under grant agreement No. 101120732 (AUTOASSESS) and the European Research Council (ERC) under grant agreement No. 864042 (AGILEFLIGHT).
    }%
}

\begin{document}

\maketitle

\begin{abstract}
\rev{A key open challenge in \revi{agile quadrotor flight} is how to combine the flexibility and task-level generality of model-free reinforcement learning (RL) with the structure and online replanning capabilities of model predictive control (MPC), aiming to leverage their complementary strengths in dynamic and uncertain environments.}
This paper provides an answer by introducing a new framework called \emph{Actor-Critic Model Predictive Control}.
The key idea is to embed a differentiable MPC within an actor-critic RL framework.
This integration allows for short-term predictive optimization of control actions through MPC, while leveraging RL for end-to-end learning and exploration over longer horizons.
Through various ablation studies, \revi{conducted in the context of agile quadrotor racing}, we expose the benefits of the proposed approach: it achieves better out-of-distribution behaviour, better robustness to changes in the \revi{quadrotor's} dynamics and improved sample efficiency.
Additionally, we conduct an empirical analysis \revi{using a quadrotor platform} that reveals a relationship between the critic's learned value function and the cost function of the differentiable MPC, providing a deeper understanding of the interplay between the critic's value and the MPC cost functions. 
Finally, we validate our method in a drone racing task on different tracks, in both simulation and the real world.
Our method achieves the same superhuman performance as state-of-the-art model-free RL, showcasing speeds of up to 21~m/s.
We show that the proposed architecture can achieve real-time control performance, learn complex behaviors via trial and error, and retain the predictive properties of the MPC to better handle out-of-distribution behavior.

\end{abstract}

\section*{Supplementary Material}
\noindent Video of the experiments: \url{https://youtu.be/_qekrF4Emzg}

\noindent Source code: \url{https://github.com/uzh-rpg/acmpc_public}

\vspace{-4pt}

\section{Introduction}
\label{sec: introduction}
\begin{figure}[htp]
\centering
\includegraphics[width=0.9\linewidth]{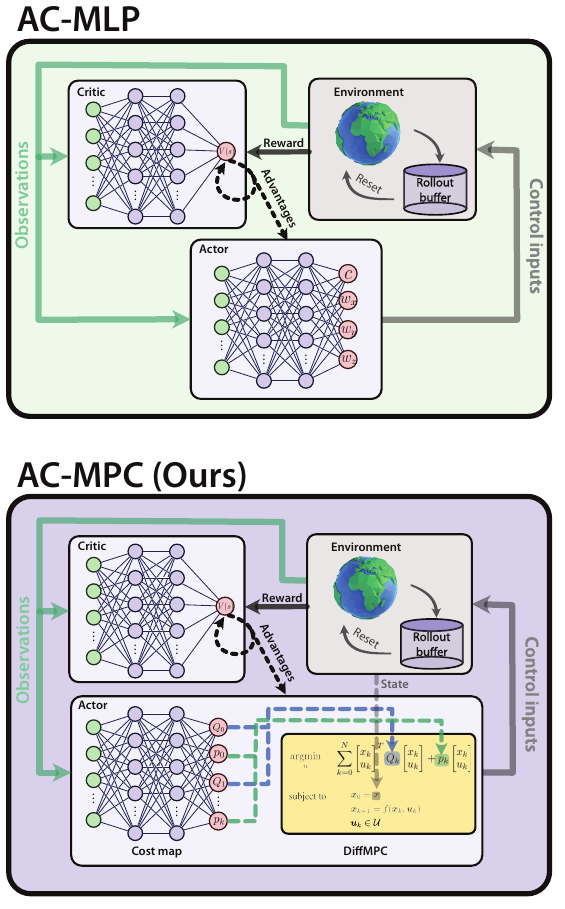}
\caption{
\textbf{Top:} A block diagram of an actor-critic reinforcement learning architecture with a Multilayer Perceptron (MLP)
\textbf{Bottom}: A block diagram of the proposed approach. 
We combine the strength of actor-critic RL and the robustness of MPC by placing a differentiable MPC as the last module of the actor policy.
At deployment time, the commands for the environment are drawn from solving an MPC, which leverages the system's dynamics and finds the optimal solution given the current state.
We show that the proposed approach achieves better out-of-distribution behavior and better robustness to changes in the dynamics. We also show that the predictions of the differentiable MPC can be used to improve the learning of the value function and the sample efficiency.
}
\label{fig:block_diagram}
\end{figure}
The animal brain's exceptional ability to quickly learn and adjust to complex behaviors is one of its most remarkable traits, which remains unattained by robotic systems.
This has often been attributed to the brain's ability to make both immediate and long-term predictions about the consequences of its actions and plan accordingly~\cite{Rao1999, Friston2010, lecun2022path}.
In robotics and control theory, model-based control has demonstrated a wide array of tasks with commendable reliability~\cite{tzanetos2022ingenuity, arthur1975applied}.
In particular, Model Predictive Control (MPC) has achieved notable success across various domains such as the operation of industrial chemical plants~\cite{ellis2016economic}, control of legged robots~\cite{wieber2016modeling}.
\revi{In this work, we focus specifically on the domain of agile drone flight, where optimization-based approaches have recently shown a series of remarkable successes~\cite{foehn2020alphapilot, foehn2021CPC, romero2022mpcc, mpcc_replanning}.}
\revi{In this context,} the effectiveness of MPC stems from its innate capability for online replanning. This enables it to make decisions that optimize a system's future states over a specified short time horizon.

However, as tasks grow in complexity, model-based approaches necessitate substantial manual engineering, tailored to each specific task. This includes the careful crafting of the cost function, \extension{tuning of \rev{parameters},} and design of an appropriate planning strategy~\cite{romero2022mpcc, tuning_effort_1}. 
Often, conservative assumptions about the task are made, leading to potentially sub-optimal task performance,
for instance, in tasks where the dynamical system is taken to its limits~\cite{foehn2021CPC, romero2022mpcc}, or in applications that require discrete mode-switching~\cite{science22hutter}.
Furthermore, the modular structure of model-based approaches may result in the progressive build-up of errors, accumulating in a cascading manner. This can compound inaccuracies, reinforce conservative estimations, and diminish the overall effectiveness of the system~\cite{Song23Reaching, Roy2021FromML, xiao2022motion}.
\rev{
Recently, a plethora of works on generative modeling approaches, particularly diffusion-based policies, have been proposed to mitigate compounding errors in sequential decision-making~\cite{reviewer4_3, reviewer4_4, reviewer4_5}.
}

\revi{These limitations have motivated a growing interest in data-driven and reinforcement-learning approaches for robotics~\cite{nature_go, mnih2015human_level, nature_starcraft, wurman2022gtsohpy}, and  specifically for human-champion performance in drone racing, both state-based~\cite{Song23Reaching} and vision-based~\cite{kaufmann23champion}, where RL controllers can be optimized directly from task-level objectives rather than handcrafted cost terms.}
\extension{
In particular, the results in~\cite{Song23Reaching} demonstrate that model-free RL can achieve superior performance compared to optimal control techniques when pushing the limits in autonomous racing. 
This success is mainly due to the ability of RL to directly optimize a non-differentiable objective, eliminating the need for proxy objectives in the form of a predefined reference time trajectory or continuous 3D path. 
More generally, RL offers a flexible approach to control by directly optimizing a feedback policy through interactions with the environment, eliminating the need for intermediate representations. Unlike trajectory optimization or standard optimal control methods, RL can handle sparse, intricate reward signals (such as minimizing crashes, energy consumption, or lap times). \rev{This combination of offline learning, combined with the use of techniques like domain randomization or curriculum learning often makes RL more scalable and effective in complex scenarios where optimal control algorithms may struggle to yield real-time optimal solutions~\cite{Song23Reaching}.}
}

However, RL architectures are not without their own set of challenges~\cite{Roy2021FromML, safe_learning_schoelig}. 
\extension{Learning a model-free, end-to-end policy without explicitly leveraging prior knowledge in the training process, such as physics or dynamic models, results in the need to learn everything from data, often resulting in millions of interactions within a simulator.}
\rev{Additionally, while RL approaches are generally better equipped to handle sparse rewards, they still struggle as reward sparsity increases~\cite{reviewer4_2, reviewer4_1}.}
While the end-to-end paradigm is attractive, it often lacks generalizability and robustness \final{to out-of-distribution scenarios}.
This has resulted in hesitancy to apply end-to-end learned architectures to safety-critical applications and has fostered the development of approaches that advocate for the introduction of safety in learned pipelines~\cite{safe_learning_schoelig, hewing2020learning, wabersich2021probabilistic}.

\extension{
This article is an extension of \cite{acmpc2024_icra}, which introduced an architecture called Actor-Critic Model Predictive Control to bridge the gap between Reinforcement Learning and Model Predictive Control, \revi{applied to agile quadrotor flight}.
This architecture equips the agent with a differentiable MPC~\cite{amos2018differentiable}, placed after the last layer of the actor network, as shown in Fig. \ref{fig:block_diagram}. This differentiable MPC module provides the system with online replanning capabilities and allows the policy to predict and optimize the short-term consequences of its actions.

Instead of relying on intermediate representations such as trajectories, we directly learn a map from observations to the cost function.
Therefore, at deployment time, the control commands are drawn from solving an MPC, which leverages the \revi{quadrotor's dynamics} and finds the optimal solution given the current state.
The differentiable MPC module, which incorporates a model of the \revi{quadrotor's dynamics}, provides the agent with prior knowledge even before any training data is received.
The second component of our actor is the cost map, a deep neural network that encapsulates the dependencies between observations and the cost function of the MPC. 
In other words, while the differentiable MPC captures temporal variations inside its horizon, the neural cost module encodes the dependencies in relation to the observations.
This architecture thereby incorporates two different time horizon scales: the MPC drives the short-term actions while the critic network manages the long-term ones.
This represents a significant advantage over vanilla actor-critic RL, where the actor is typically a randomly initialized feedforward neural network with no domain-specific structure nor priors. 

\cite{acmpc2024_icra} demonstrated that the AC-MPC architecture exhibits more stable behavior in out-of-distribution scenarios \revi{in agile quadrotor flight tasks}, highlighting its generalizability compared to its neural-network-only counterpart. 
Lastly,~\cite{acmpc2024_icra} showcased the applicability of the proposed approach in real-world \revi{drone racing}, demonstrating the feasibility and effectiveness of the AC-MPC architecture.

This paper extends~\cite{acmpc2024_icra} as follows:
\rev{
\begin{itemize}
    \item \textbf{Model Predictive Value Expansion (MPVE).}  
    We introduce an extension to our framework that incorporates Model Predictive Value Expansion, a critic-training scheme that re-uses the short-horizon state-action predictions produced by the differentiable MPC during every forward pass. 
    Unlike general value expansion methods that may require a separate policy for generating rollouts, our approach efficiently improves the critic without incurring additional policy call overhead. 
    Our analysis, \revi{performed on quadrotor systems}, demonstrates that this method improves sample efficiency, with the benefits being most pronounced for small values of the TD parameter $\lambda$.

     \item \textbf{Relationship between value function and MPC cost.}  
          We perform a systematic analysis which reveals that the Hessian of the learned value function closely matches the quadratic cost matrices generated by the neural cost map.  
          \revi{This empirical finding, obtained for the task of stabilizing a quadrotor system around hover,} establishes a previously unknown fact, that there is a concrete relationship between what the critic learns and the optimisation landscape of the solver.
          This result not only provides a deeper understanding of the interplay between the RL value and the MPC cost functions within the AC-MPC framework, but also opens a path toward a deeper understanding of this type of controllers.

    \item \textbf{Exploration analysis} We show that after optimizing hyperparameters, specifically exploration, AC-MPC can leverage the prior knowledge included in the \revi{quadrotor dynamics} to achieve better training performance than AC-MLP \revi{on the drone racing task}.
    \item \textbf{A broad set of ablations and additional comparisons:} 
    (i) we conduct an experiment where we show that, \revi{for the task of flying a quadrotor through a race track}, AC-MPC is more robust to dynamic parameter changes than the AC-MLP baseline without the need of retraining the policy, keeping high success rates for variations in \revi{the quadrotor's mass, XY inertia, and body rate limits}. This improved robustness is attributed to the differentiable MPC's access to the \revi{quadrotor's dynamics}. These results highlight AC‑MPC’s ability to cope with dynamics uncertainty, which is of paramount importance when testing in the extremely challenging real-world \revi{drone racing} benchmark.
    (ii) we conduct an extra empirical analysis where we compare the \revi{training} performance of our architecture \revi{in the drone racing task} for the choice of three different matrix representations: \emph{Diagonal}, \emph{Cholesky} and \emph{Full Matrix}.
    (iii) we have added an additional comparison with an adaptive state-of-the-art L1-MPC controller.
    This comparison with a robust controller strategy in the cases where we show the benefits of our architecture to out of distribution behaviour helps understanding the difficulty of the task and puts our results into a new perspective.
    \item \textbf{Extended real-world validation}. We demonstrate in both simulation and real-world experiments that the proposed AC-MPC achieves superhuman performance in the extremely challenging task of drone racing, attaining speeds of up to 21~m/s on a real quadrotor platform.
    These results are on par with state-of-the-art model-free reinforcement learning and highlight that the strengths of AC-MPC do not come at the expense of performance.
    \item \textbf{Source code}. We open source the AC-MPC module: \url{https://github.com/uzh-rpg/acmpc_public}
\end{itemize}
}
The AC-MPC architecture not only advances the integration of RL and MPC but also offers practical improvements in out-of-distribution behavior, training efficiency, and real-world applicability \revi{in the context of agile quadrotor flight}, contributing to the development of more robust and efficient control systems.

}

\section{Related Work}
\label{sec: related work}
\extension{
\textbf{Model Predictive Control (MPC) in robotics}. 
MPC has had an enormous success in robotic applications over the last decades~\cite{garcia1989model, wang2009fast, mpcoffroad, liniger2015optimization, foehn2021CPC, romero2022mpcc, mpcc_replanning, farshidian2017real, liu2017mpcav, han2023modelaggressivedriving, arcari2023bayesian, saviolo2023activempc, li2023nonlinearmpc, frohlich2022contextual, krinner2024mpcc++, safetyfilter, tagliabue2024efficient, tagliabue2024tube, djeumouone}. 
However, often the cost function needs to be crafted by hand by an expert, and the hyperparameters need to be tuned at deployment time~\cite{tuning_effort_2, tuning_effort_3}.
Additionally, in most MPC setups~\cite{bledt2018cheetah, neunert2018whole, neunert2016fast, wieber2016modeling, bjelonic2022offline, kuindersma2016optimization, foehn2021CPC}, the high-level task is first converted into a reference trajectory (planning) and then tracked by a controller (control).
This decomposition of the problem into these distinct layers is greatly favored by the MPC methodology, largely due to the differentiability and continuity requirement of optimal control's cost functions.
The cost function is shaped to achieve accurate trajectory tracking and is decoupled from (and usually unrelated to) the high-level task objective.
As a result, the hierarchical separation of the information between two components leads to systems that can become erratic in the presence of unmodeled dynamics.
In practice, a series of conservative assumptions or approximations are required to counteract model mismatches and maintain controllability, resulting in \rev{controllers} that are no longer optimal \cite{Song23Reaching, Roy2021FromML, xiao2022motion}. 

Another promising model-based direction for addressing complex control challenges is sampling-based MPC algorithms~\cite{williams2018information, bhardwaj2022stormbyronboots, minarik2024mppirobert, xue2024fullmppidifussion}, which are designed to handle intricate, non-differentiable cost functions and general nonlinear dynamics. 
These algorithms integrate system dynamics -- either known or learned -- into the Model Predictive Path Integral (MPPI) control framework, optimizing control in real-time.
A key feature of sampling-based MPC is the ability to generate a large number of control samples on-the-fly, often leveraging parallel computation via Graphics Processing Units (GPUs). 
However, implementing sampling-based MPC on embedded systems poses significant challenges, as it tends to be both computationally demanding and memory-intensive.

\textbf{Reinforcement Learning (RL) in robotics}.
RL has risen as an attractive alternative to conventional controller design, achieving impressive performance that goes beyond model-based control in a variety of domains~\cite{nature_go, andrychowicz2020learning, nature_starcraft, nature_rl_fusion, nature_sorting, science20hutter}.
RL optimizes a controller using sampled data and rev{can easily handle non-differentiable models and sparse objectives}, manifesting great flexibility in controller design. Compared to model-based optimal control, RL has a number of key advantages: most importantly, RL can optimize the performance objective of the task directly, removing the need for explicit intermediate representations, such as trajectories. Moreover, RL can solve complex tasks from raw, high-dimensional sensory input, without the need of a specific metric state~\cite{Song23Reaching}.
However, RL is still far from reaching the level of robustness and generalisability of model-based control approaches when deployed in the real world, due to its brittleness when deployed in situations outside of the training distribution, and its lack of guarantees~\cite{safe_learning_schoelig}.
Model-free methods generally use black-box optimization methods and do not exploit the first-order gradient through the dynamics, thus cannot leverage the full advantage of the prior knowledge.

\textbf{Combinations of MPC and learning}.
Several methods have been developed to learn cost functions for MPC~\cite{BayesianOptimizationECC, wml, romero2019nonlinear, tearle2021predictive, grandesso2023cacto, hoeller2020deepvalue, reiter2024ac4mpc, sacks2022learning, sacks2024deep, zarrouki2024safeWVMPC, zarrouki2021weightsWVMPC, jenelten2023dtc, reiter2023ecc, seel2022convex}, dynamics models for MPC~\cite{frohlich2022contextual, saviolo2022physics, deepmpc, watter2015embed, ebert2018visual, reviewer5_1, reviewer5_2}, or both simultaneously~\cite{hansen2022temporal, krinner2024mpcc++, amos2018differentiable, pereira2018mpc, adhau2024reinforcement, bohn2023optimization, esfahani2023learning, anand2023painless, cai2023learning, reviewer1_1, reviewer1_2}.
For example, in~\cite{wml}, a policy search strategy is adopted that allows for learning the hyperparameters of a loss function for complex agile flight tasks. On the other hand, the works in~\cite{frohlich2022contextual, BayesianOptimizationECC} use Bayesian Optimization to tune the hyperparameters and dynamics of MPC controllers for different tasks such as car racing.
Similarly, in \cite{sacks2024deep} the authors propose a new method called \emph{Deep Model Predictive Optimization} (DMPO) that learns the update rule of an MPPI controller using RL. They evaluate their algorihtm on a real quadrotor platform and outperform state-of-the-art MPPI algorithms.

Recent model-based RL approaches integrate MPC-like components either during policy learning~\cite{hafnerdream, aljalbout2024limt} or by using explicit MPC-based policies~\cite{nagabandi2018neural,chua2018deep,hansen2022temporal, hansen2024tdmpc2}.
Many of these methods learn a dynamics model and use it to perform MPC at inference time.
The latter step is either done via random shooting~\cite{nagabandi2018neural}, the cross-entropy method~\cite{chua2018deep, nagabandi2020deep}, or a combination of these methods with a learned critic~\cite{sikchi2022learning, hansen2022temporal}.

Alternatively, approaches leveraging differentiability through optimizers have been on the rise. For example, for tuning linear controllers by getting the analytic gradients~\cite{difftune}, for differentiating through an optimization problem for planning the trajectory for a legged robot~\cite{yang2023iplanner}, or  for creating a differentiable prediction, planning and controller pipeline for autonomous vehicles~\cite{karkus2023diffstack}, or for moving horizon estimation \cite{wang2023neuralmhe}.

On this same direction, MPC with differentiable optimization~\cite{amos2018differentiable, theseus, pypose, infinitediffmpc} proposed to learn the cost or dynamics of a controller end-to-end. 
In particular, the authors in~\cite{amos2018differentiable} were able to recover the tuning hyperparameters of an MPC via imitation learning for non-linear, low-dimensional dynamics -- such as cartpole and inverted pendulum -- by backpropagating through the MPC itself.
Later, in~\cite{google_performer}, the authors augment the cost function of a nominal MPC with a learned cost that uses the gradient through the optimizer for the task of navigating around humans.
To train this learned cost, they also learn from demonstrations.
Until \cite{acmpc2024_icra}, all these approaches were only demonstrated in the context of imitation learning. While imitation learning is effective, its heavy reliance on expert demonstrations is a burden. This dependence prevents exploration, potentially inhibiting its broader capabilities.
}

In \cite {acmpc2024_icra}, we address this issue by leveraging the advantages of both differentiable MPC and model-free reinforcement learning. 
By equipping the actor with a differentiable MPC, our approach provides the agent with online replanning capabilities and with prior knowledge, which is a significant advantage over model-free RL, where the actor is a randomly initialized feedforward neural network. \rev{Unlike conventional MPC, our approach benefits from RL training techniques like domain randomization, flexibly allowing for the optimization of intricate objectives through iterative exploration and refinement.}
\rev{

\textbf{Real-World Drone Racing as a benchmark.}
Controlling a real world robot at the limits of handling is extremely hard, more so when dealing with an inherently unstable system such as a quadrotor.
In order to show how challenging real-world drone racing is, in this paragraph we cronologically go through the developments in autonomous drone racing over the years.
We start with polynomial trajectories being tracked by a tracking MPC controller, in the AlphaPilot \cite{croonalphapilot, foehn2020alphapilot} competition. 
Both winning and second place used classical methods to generate polynomial minimum snap trajectories that could be tracked using a geometric or a MPC controller.
However, these trajectories didn't result in time optimal performance, since they were not driving the actuators to saturation.
After this and because of this lack of optimality, the first time-optimal flight planner was developed \cite{foehn2021CPC}. 
However, the trajectories that were generated using this planner, although they resulted in the lowest time possible, couldn't be tracked in the real hardware because the commands would be in saturation almost all of the time, and any model mistmatch, small disturbance, or state estimation error would drive the system to crash due to the lack of control authority in saturation.
To allow control authority to be able to deploy the system, the authors needed to be extremely conservative with the planner, reducing the maximum thrust that the quadrotor was able to generate by up to 30\%, resulting in final, real-world lap times that were not competitive with best human pilots.
As a response to this, a new model-based approach, Model Predictive Contouring Control \cite{romero2022mpcc, krinner2024mpcc++} was developed, which was able to achieve better lap times than previous approaches.
However, this approach needed to be tuned by a human in the real-world, resulting in extremely long tuning times, including tuning of parameters that govern the trade-off between path progress maximization and contouring error minimization. 
These parameters need to be tuned in the physical world for every different track. 
This approach became the state of the art of model-based drone racing, after many years of development.
In a follow up publication ~\cite{Song23Reaching}, the authors extensively compared the best MPC and RL approaches for drone racing, showing that RL approach can achieve superhuman lap times with high success rate in the real world and with significantly less engineering effort.

The history of the field demonstrates that achieving robust, superhuman performance in drone racing requires overcoming significant real-world robotic challenges.
Therefore, we believe drone racing stands as a deeply challenging benchmark, perfectly suited for validating advanced control algorithms.

Our publication aims to close the gap and show that when using our architecture, an MPC controller can be synthetized that achieves the same performance, same success rates in the real world and better out-of-distribution behaviour than an RL controller in this extremely challenging benchmark.

}

\section{Methodology}
\label{sec: methodology}
\rev{In this section, we first present preliminaries about model-based control and reinforcement learning, and then introduce our method.}
\subsection{Preliminaries}
Consider the discrete-time dynamic syste with continuous state and input spaces, $\vec{x}_k \in \mathcal{X}$ and $\vec{u}_k \in \mathcal{U}$ respectively. 
Let us denote the time discretized evolution of the system \mbox{$f : \mathcal{X} \times \mathcal{U} \mapsto \mathcal{X}$} such that $\vec{x}_{k + 1} = f(\vec{x}_k, \vec{u}_k)$, where the sub-index $k$ is used to denote states and inputs at time $t_k$.
\rev{We consider the general task of finding a control policy $\pi(\vec{x})$, which maps the current state to the optimal input, $\pi : \mathcal{X} \mapsto \mathcal{U}$, such that the cost function $J: \mathcal{X} \mapsto \mathbb{R}^+$ is minimized,}
\begin{align}
\pi(\vec{x}) = & \argmin_{\vec{u}}
  & & 
    J(\vec{x}) \notag\\
        & \text{subject to} && \vec{x}_0 = \vec{x}, \quad \vec{x}_{k+1} = f(\vec{x}_k, \vec{u}_k) \notag\\
        &&& \final{\vec{u}_k \in \mathcal{U}},
        \label{eq:ocp}
\end{align}
where $k$ ranges from $0$ to $N$ for $x_k$ and from $0$ to $N-1$ for $\vec{u}_k$.
\extension{
In the following sections, we will go through different ways in which $J(\vec{x})$ can be chosen.
}

\extension{
\subsection{Tracking Model Predictive Control}
\label{sec:mpc}
One of the primary uses of MPC is to control a dynamical system's behavior such that it closely follows a pre-computed reference trajectory. 
This is known as trajectory tracking, and the corresponding MPC formulation is referred to as \emph{tracking MPC}.
In tracking MPC approaches~\cite{MPCSurveyASL_2020, Bangura14ifac, Diehl2006springer}, the objective $J(\vec{x})$ is to minimize a quadratic penalty on the error between the predicted states and inputs, and a given dynamically feasible reference $\vec{x}_{k,ref}$ and $\vec{u}_{k, ref}$. 
Consequently, the cost function $J(\vec{x})$ in problem \eqref{eq:ocp} is substituted by:
\begin{align}
    \label{eq:J_mpc}
    J_{MPC}(\vec{x}) = \sum_{k = 0}^{N-1} \Vert \Delta \vec{x}_k \Vert_{\bm{Q}}^2 + \Vert \Delta \vec{u}_k \Vert_{\bm{R}}^2 + \Vert \Delta \vec{x}_N \Vert_{\bm{P}}^2,
\end{align}
where $\Delta \vec{x}_k = \vec{x}_k - \vec{x}_{k,ref}$, $\Delta \vec{u}_k = \vec{u}_k - \vec{u}_{k,ref}$, and where $\vec{Q} \succeq 0$, $\vec{R} \succ 0$ and $\vec{P} \succeq 0$ are the state, input, and final state weighting matrices. 
The norms of the form $\Vert \cdot \Vert_A^2$ represent the weighted Euclidean inner product $\Vert \vec{v} \Vert^2_A = \vec{v}^T A \vec{v}$. 

This formulation relies on the fact that a feasible reference $\vec{x}_{k,ref}$, $\vec{u}_{k,ref}$ is accessible for $N$ time steps in the horizon. 
Searching for this reference trajectory is often referred to as \emph{planning}, which, depending on the application, can be both computationally intensive and complex, particularly in scenarios involving complex dynamics or cluttered environments~\cite{foehn2021CPC, Song_Steinweg_Kaufmann_Scaramuzza_2021, penicka2022cluttered}. In many cases, solving the planning problem becomes the bottleneck in real-time applications due to high computational requirements and the need for precise system models.

The tracking MPC formulation also relies on the assumption that a trajectory represents a well-posed, quadratic proxy objective for solving the end task.
While this assumption holds for many practical use cases, it becomes limiting in situations where the system operates near its performance limits, such as aggressive maneuvers or high-speed tasks, where the trajectory may no longer serve as an effective proxy. 
\rev{In such cases, tracking a optimal reference in the real world can result in suboptimal performance or even task failure~\cite{Song23Reaching}.}

\subsection{Economic Model Predictive Control}
\label{sec:empc}
To overcome these limitations, Economic MPC directly incorporates the task objective into the optimization problem, eliminating the need for a pre-defined reference trajectory. 
This fundamental difference allows for a more flexible cost function design, enabling the controller to be tailored to the specific economic objectives of the task \cite{ellis2014tutorialeconomic, rawlings2012fundamentalseconomic}:
\begin{align}
    \label{eq:J_empc}
    J_{EMPC}(\vec{x}) = \sum_{k = 0}^{N-1} l(x_k, u_k),
\end{align}
where $l(x_k, u_k)$ is directly the stage cost.
This stage cost is generally designed to optimize sparse metrics, such as energy consumption, efficiency, or success rate.

While economic MPC offers advantages in terms of flexibility with respect to task design, it also presents several challenges that must be addressed for successful implementation. 
One of the primary caveats is that the stage cost $l(x_k, u_k)$ in most tasks tends to be sparse and non-differentiable, which complicates the formulation of Eq. \eqref{eq:J_empc}.
For example, in tasks with discontinuous behavior (e.g., switching between different modes, dealing with contact forces, or dealing with agent crashes), it is generally difficult to find a cost function that is differentiable.
}

\extension{
\subsection{General Quadratic MPC formulation}
In most MPC approaches, there is a need of i) an explicit manual selection of a differentiable cost function that properly encodes the end task, and ii) hyperparameter tuning. 
As mentioned in section \ref{sec:mpc}, in the case of a standard tracking MPC this encoding is done through planning, by finding a dynamically feasible reference trajectory that translates the task into suitable cost function coefficients for every time step.
For economic MPC (section \ref{sec:empc}), recent research has been going in the direction of using data to approximate the cost function~\cite{gros2019datadriveneconomic}, or making it computationally tractable by transforming it to a quadratic program using the Hessian~\cite{verschueren2016economic, zanon2020gausseconomic}.

However, these approaches present two main drawbacks: i) hand-crafting a dense, differentiable cost function can be difficult for a general task, and ii) even if this cost function is found, extra effort needs to be spent in fine tuning the hyperparameters for real-world deployment.
More generally, optimization-based architectures such as MPC need to run in real-time when deployment in the real world is desired.
As a result, the underlying optimization problem is often simplified by approximating the original nonlinear formulation and converting it into a Quadratic Program (QP) suitable for the real-time iteration (RTI)~\cite{diehl2005RTI, gros2020linearRTI, Verschueren2021} scheme to ensure computational tractability.
A general quadratic cost function can be written as 
\vspace{-4pt}
\begin{align}
    \label{eq:J_general_mpc}
    J_{QP}(\vec{x}) = \sum_{k = 0}^{N} \mat{x_k \\ u_k}^T Q_k \mat{x_k  \\ u_k} + p_k \mat{x_k \\ u_k}.
\end{align}
\rev{
Instead of manually designing a complex, non-linear cost function that would subsequently require approximation, we propose to learn the final quadratic approximation directly.
In this paper, we propose to directly search for the matrix coefficients of Eq. \eqref{eq:J_general_mpc} (i.e.,  $Q_k$ and $p_k$) using reinforcement learning and differentiable MPC~\cite{amos2018differentiable}.
This approach allows the reinforcement learning agent to discover the most effective quadratic representation of the high-level task objective.
By doing so, we can encode the task using a general, non-differentiable reward function while simultaneously producing a controller that is already computationally tractable, ready for real-time deployment without further approximations.
}

}
\subsection{Actor-Critic Reinforcement Learning}
\extension{
Reinforcement learning problems are often framed within the Markov Decision Process (MDP) formalism, which provides a structured approach for modeling decision-making problems. An MDP is defined by the tuple $(\mathcal{S}, \mathcal{A}, P, R, \gamma)$, where $\mathcal{S}$ is the set of possible states, $\mathcal{A}$ is the set of possible actions, $P(s_{k+1} | s_k, a_k)$ is the state transition probability, $R(s_k, a_k)$ is the reward function, and $\gamma \in [0, 1)$ is the discount factor. The objective in an MDP is to find a policy $\pi$ that maximizes the expected cumulative reward over time.
Within the context of Problem \eqref{eq:ocp}, and when the state transition function is deterministic, it is equivalent to the system dynamics $f(x_k, u_k)$, where in this case the observation $s_k$ is directly the state $x_k$ and the action $a_k$ is directly the control input $u_k$.
}

\extension{
To solve such problems, a common approach in Reinforcement Learning is to optimize directly the policy $\pi_{\theta}$ by using the policy gradient. The policy  $\pi_{\theta}$ is parameterized by $\theta$, typically the weights of a neural network. The goal is to adjust these parameters to maximize the expected return, defined in the infinite-horizon case as:
\begin{align} R(\tau) = \sum_{k=0}^{\infty} \gamma^{k} r(s_k, a_k), \end{align}
where $\tau$ denotes a trajectory, $r(s_k, a_k)$ the reward function, and $\gamma \in [0, 1)$ is a discount factor that discounts future rewards.
\rev{Unlike real-time optimization methods like MPC, RL can handle non-differentiable, non-derivable models and easily optimize sparse rewards.}
Additionally, policy gradient optimization is typically performed offline through interactions with simulation.
Once trained, the policy $\pi_{\theta}$ enables the computation of control signals by simply evaluating the function $a^{\ast} = \pi_{\theta}(s_k)$ at each time step, reducing the computational complexity during deployment.
}
For Actor-Critic reinforcement learning, the key idea is to simultaneously learn a state-value function $V_{\omega}(s)$ and learn a policy function $\pi_{\theta}$, where the value function (Critic) and policy (Actor) are parameterized by $\omega$ and $\theta$ separately. 
\rev{The policy is updated via policy gradient~\cite{sutton1999policy},}
\begin{align}
    \label{eq:policy_gradient}
    \nabla_{\theta}J(\pi_{\theta}) = \frac{1}{N} \sum_{i=1}^N \sum_{k=1}^T \nabla_{\theta} \log \pi_{\theta}(a_k^i | s_k^i) A_{\omega}(s_k^i, a_k^i),
\end{align}
where $A(s_k, a_k) = r(s_k, a_k) + \gamma V_{\omega}(s_{k+1}) - V_{\omega}(s_k)$ is the advantage function used as a baseline. In a standard actor-critic method, the policy is a stochastic representation where $a_k=\pi_{\theta}(s_k)$ is the mean of a Gaussian distribution.
\subsection{Actor-Critic Model Predictive Control}
\label{sec:architecture}
This paper proposes an Actor-Critic MPC controller architecture where the MPC is differentiable~\cite{amos2018differentiable} and the cost function is learned end-to-end using RL. 
The MPC block is introduced as the differentiable module of the actor in an actor-critic pipeline that uses the proximal policy optimization~(PPO) algorithm~\cite{schulman2017proximal}. Our architecture is shown in Fig.~\ref{fig:block_diagram}.
In contrast to previous work~\cite{wml}, where the MPC is taken as a black-box controller and the gradient is sampled, in our case the gradient of the cost function with respect to the solution is analytically computed and propagated using a differentiable MPC~\cite{amos2018differentiable}. Therefore, for every backward and forward pass of the actor network, we need to solve an optimization problem.
Instead of resorting to task-specific engineering of the cost function, we propose a neural cost map where the $Q_k$ and $p_k$ terms are the output of the neural network layers preceding the differentiable MPC block.
This allows us to encode the end task directly as a reward function, which is then trainable end-to-end using the PPO training scheme.
The main benefit of this approach with respect to training a standard Multi-Layer Perceptron (MLP) end-to-end is that the final module of the actor is a model-based MPC controller,
\begin{align}
    \label{eq:diffMPC}
    u_k \sim \mathcal{N} \{\text{diffMPC}(x_k, Q(s_k), p(s_k) ), \Sigma \},
\end{align}
and therefore it retains its \rev{generalizability to out-of-distribution situations}.
The model-based controller placed after the last layer of the actor network ensures that the commands are always feasible for the dynamics at hand, and that they respect the system constraints.
This differentiable MPC supports control input constraints but not state constraints, which is why  we add $\vec{u}_k \in \mathcal{U}$ in \eqref{eq:ocp}.
To allow for exploration, during training the control inputs are sampled from a Gaussian distribution where the mean is the output of the MPC block, and the variance is controlled by the PPO algorithm.
However, during deployment the output from the MPC is used directly on the system without further sampling, retaining all properties of a model-based controller.
\rev{
Algorithm \ref{algo: ac-mpc} shows a high-level description of the AC-MPC algorithm.
}
{\scriptsize
\begin{algorithm}[!htp]
    \small
    \caption{{\bf \small Actor-Critic Model Predictive Control}   
    \label{algo: ac-mpc}}
    \KwIn{ initial neural cost map, initial value function $V$}
    \textbf{for} $i=0, 1, 2, \cdots $ \textbf{do} \\
    \quad Collect set of trajectories $\mathcal{D}_i\{\tau\}$ with \\ \quad\quad$u_k \sim  \mathcal{N} \{\text{diffMPC}(x_k, Q(s_k), p(s_k) ), \Sigma \} $ \\
    \quad Compute reward-to-go $\hat{R}_k$ \\
    \quad Compute advantage estimates $\hat{A}_k$ based on value \\
    \quad\quad function $V(s_k)$\\
    \quad  Update the cost map by policy gradient (e.g., PPO-clip\\
    \quad\quad objective) and diffMPC backward~\cite{amos2018differentiable} \\
    \quad  Fit value function by regression on mean-squared error \\
    \KwOut{Learned cost map}
\end{algorithm}
}

\subsection{Neural Cost Map}
The cost function for the model predictive control architecture presented in Section \ref{sec:architecture} is learned as a neural network, depicted in Fig. \ref{fig:block_diagram} as \emph{Cost Map}.
Several adaptations to the system are needed in order to properly interface the neural network architecture with the optimization problem.
First, we construct the matrix $Q(s_k)$ and vector $p(s_k)$ as follows:
\begin{align}
    Q(s_k) &= \text{diag}(Q(s_k)_{x_1}, \dots, R(s_k)_{u_1}, \dots) \notag \\ 
    p(s_k) &= [p(s_k)_{x_1}, \dots, p(s_k)_{u_1}, \dots] \quad\quad\forall k \in 0, \dots, T ,\notag
\end{align}
where $x_1, \dots$ and $u_1, \dots $ are the states and inputs to the system, respectively, and $Q(s_k)$ and $p(s_k)$ are the learnable parameters, interface from the neural network to the optimization problem.

The purpose of the diagonalization of the $Q$ matrix is to reduce the dimensionality of the learnable parameter space. 
Therefore, the dimensionality of the output dimension of the \emph{Cost Map} is $2T(n_{state} + n_{input})$.
In order to ensure the positive semi-definiteness of the $Q$ matrix and the positive definiteness of the R matrix, a lower bound on the value of these coefficients needs to be set.
To this end, the last layer of the neural cost map has been chosen to be a sigmoid which allows for upper and lower bounds on the output value.
These lower and upper limits are chosen equal for $Q$ and $p$, of 0.1 and 100000.0, respectively.
\final{Therefore}, the final neural cost map consists of two hidden layers of width 512 with ReLUs in between and a sigmoid non-linearity at the end.
The critic network consists also of two hidden layers of width 512 and ReLUs. The output of the critic network is a scalar.

\extension{
\subsection{Model-Predictive Value Expansion}
\label{sec:mpve}
At each control iteration, the differentiable MPC block in Fig. \ref{fig:block_diagram} outputs a sequence of optimal states and actions, of which only the first action $u_0$ is applied to the system.
The remaining predicted states and actions, although currently discarded, contain valuable information. They can be used to improve the learning of the value function.

In this section, we propose an extension to our algorithm that makes use of the predictions of the MPC to improve the quality of the value function estimate. We call this extension Model-Predictive Value Expansion (MPVE).
This idea is inspired by the Model-Based Value Expansion algorithm proposed by~\cite{feinberg2018model}, which uses the extra predicted rollouts using a learned model to improve the value function estimate.
Assuming that our model is \rev{accurate up to $H$ horizon steps} and that we have access to the reward of the predictions, one can define the H-Step Model-Predictive Value Expansion as
\begin{align}
  \hat{V}_H(s) = \sum_{t=0}^{H-1} \gamma^t \hat{r} + \gamma^{H} \hat{V}(s_{H}),
\end{align}
where $\hat{r}$ represents the predicted rewards, and $\hat{V}(s_{H})$ is the estimated value of the predicted state at the end of the horizon $H$.
  
However, as explained in~\cite{feinberg2018model}, a challenge that arises is the distribution mismatch. This occurs when the distribution of states seen during training differs from the distribution of states encountered when using the predictions from the differentiable MPC. This mismatch can degrade performance, as highlighted in~\cite{feinberg2018model}.

To address this issue, the work in~\cite{feinberg2018model} incorporates the TD k-trick, an approach designed to mitigate the distribution mismatch problem by aligning the training distribution with the prediction's distribution over multiple steps. The TD k-trick involves training the value function on k-step returns, which helps ensure the training and prediction distributions are closely aligned.

Therefore the expression for the value loss is extended
\begin{align}
\label{eq:pve}
  \frac{1}{H} \sum_{t=0}^{H-1} \left( V(\hat{s}_t) - \left( \sum_{k = t}^{H-1} \gamma^{k-t} \hat{r}_k + \gamma^{H} \hat{V}(\hat{s}_H) \right) \right)^2.
\end{align}
This additional term encourages the value function to be consistent with the MPC predictions across multiple time steps, thereby improving the overall accuracy of the value estimates.

We incorporate Eq. \eqref{eq:pve} into our original algorithm (Algorithm \ref{algo: ac-mpc}), to create a more robust learning process that effectively utilizes the predictive power of the MPC. This integration allows for better exploitation of the model's predictions, potentially leading to faster convergence and improved policy performance.
After including Eq. \eqref{eq:pve} in algorithm \ref{algo: ac-mpc}, this is how the algorithm changes:

{\scriptsize
\begin{algorithm}[!htp]
    \small
    \caption{{\bf \small Actor-Critic Model Predictive Control with Model-Predictive Value Expansion}
    \label{algo: ac-mpc-pve}}
    \KwIn{ initial neural cost map, initial value function $V$}
    \textbf{for} $i=0, 1, 2, \cdots $ \textbf{do} \\
    \quad Collect set of trajectories and predictions $\mathcal{D}_i\{\tau\}$ with \quad\quad$x_{k:k+H}, u_{k:k+H} \sim  \mathcal{N} \{\text{diffMPC}(x_k, Q(s_k), p(s_k) ), \Sigma \} $ \\
    \quad Compute reward-to-go $\hat{R}_k$ and value targets using TD($\lambda$)\\
    \quad Compute advantage estimates $\hat{A}_k$ based on value \\
        \quad\quad function $V(s_k)$\\
    \quad Compute reward-to-go for predictions $\hat{R}_{k:k+H}$ and \\ \quad\quad value targets using TD k-trick \\
    \quad  Update the cost map by policy gradient (e.g., PPO-clip\\
    \quad\quad objective) and diffMPC backward~\cite{amos2018differentiable} \\
    \quad  Fit value function by regression on mean-squared error \\ \quad\quad using a sum of the TD($\lambda$) value loss and Eq. \eqref{eq:pve} \\
    \KwOut{Learned cost map}
\end{algorithm}

}

}

\begin{figure}[tp]
\centering
\includegraphics[width=1.0\linewidth]{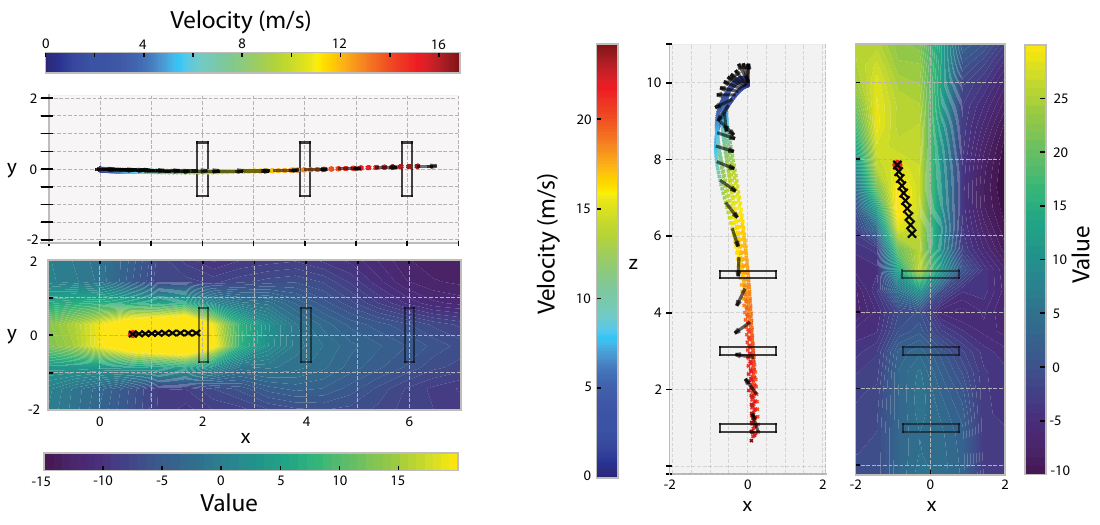}
\caption{
\textbf{Actor-Critic Model Predictive Control (AC-MPC) applied to agile \revi{quadrotor} flight:} velocity profiles and corresponding value function plots. 
The left side illustrates horizontal flight, while the right side shows vertical flight.
In the value function plots, areas with high values (depicted in yellow) indicate regions with the highest expected returns.
The MPC predictions are shown as black Xs.
}
\label{fig:horizontal_vertical_value_function}
\end{figure}
\extension{
\section{Experimental Setup}
}
\label{sec: exp}
This section presents a set of experiments, both in simulation and the real world.
\revi{All experiments in this work are conducted on a quadrotor system, both in simulation and on a real platform.}
To showcase the capabilities of our method, we have chosen the task of agile flight through a series of gates in different configurations: horizontal, vertical, circular, and SplitS. Additionally, to show the sim-to-real transfer capabilities, both circular and SplitS tracks are deployed in the real world.
We train in a simple simulator in order to speed up the training and evaluate in BEM, a high-fidelity simulator~\cite{bem}, which has a higher level of similarity in terms of aerodynamics with the real world.
The quadrotor platform's dynamics are the same as in~\cite{Song23Reaching}.
For every different task, the policies are retrained from scratch.
Lastly, all experiments have been conducted using a modification of the \emph{Flightmare} software package~\cite{song2020flightmare} for the quadrotor environment and PPO implementation and \emph{Agilicious}~\cite{agilicious} for the simulation and deployment.
\extension{

}
\subsection{Observations, actions and rewards}
\label{sec: observations_inputs_rewards}
\subsubsection{Observation space}
\revi{All tasks presented in our manuscript are quadrotor tasks, and the observation space has been tailored for such. The observation space for the quadrotor does not change, and consists of two main parts: the quadrotor's state observation~$\mathbf{o}_t^\text{quad}$ and the race track observation~$\mathbf{o}_t^\text{track}$. 
We define the quadrotor's state observation as ${\mathbf{o}_{t}^\text{quad} = [\vec{v}_{t}, \mathbf{R}_{t}] \in \mathbb{R}^{12}}$, which corresponds to the quadrotor's linear velocity and rotation matrix. 
}
We define the track observation vector as $\mathbf{o}_{t}^\text{track}=[ \delta \vec{p}_1, \cdots,  \delta\vec{p}_i , \cdots], \; i \in [1, \cdots, \rev{G}]$,
where $ \delta\vec{p}_i  \in \mathbb{R}^{12} $ denotes the relative position between the vehicle center and the four corners of the next target gate~$i$ or the relative difference in corner distance between two consecutive gates. 
Here $\rev{G}\in \mathbb{Z}^+$ represents the total number of future gates. 
This formulation of the track observation allows us to incorporate an arbitrary number of future gates into the observation. 
We use $\rev{G}=2$, meaning that we observe the four corners of the next two target gates. 
We normalize the observation by calculating the mean and standard deviation of the input observations at each training iteration.
\subsubsection{Action space}
Previous work has demonstrated the high importance of the action space choice for learning and sim-to-real transfer of robot control policies~\cite{kaufmann2022benchmark, aljalbout2024role}.
\revi{In our work, the quadrotor control input modality} is collective thrust and body rates, which was previously shown to perform best for agile flight~\cite{kaufmann2022benchmark}.
This action is expressed as a 4-dimensional vector $\mathbf{a} = [c, \omega_x, \omega_y, \omega_z ] \in \mathbb{R}^4$, representing mass-normalized thrust and body rates, in each axis separately. %
Even if the MPC block uses a model that limits the actuation at the single rotor thrust level, collective thrust and body rates are computed from these and applied to the system.
This ensures that the computed inputs are feasible for the model of the platform.
\subsubsection{Rewards}
\label{sec:rewards}
\revi{For all our experiments, the reward is tailored for quadrotor racing through gates. As such, one reward term in common is the gate progress reward, which encourages fast flight through the track.}
The objective is to directly maximize progress toward the center of the next gate. 
Once the current gate is passed, the target gate switches to the next one.
At each simulation time step $k$, the reward function is defined by 
\extension{
\begin{small}
\begin{equation}
r(k) = 
\begin{cases} 
-10.0 & \text{if collision}, \\
+10.0 & \text{if gate passed,} \\
+10.0 & \text{if race finished}, \\
\Vert g_k - p_{k-1} \Vert - \Vert g_k - p_k \Vert - b \| \boldsymbol{\omega}_k \| & \text{otherwise}.
\end{cases}
\label{eq: gate_progress}
\end{equation}
\end{small}
}
where $g_k$ represents the target gate center, and $p_k$ and $p_{k-1}$ are the vehicle positions at the current and previous time steps, respectively. Here, $ b \|\boldsymbol{\omega}_k \|$ is a penalty on the body rate multiplied by a coefficient $b=0.01$.

\extension{
\section{Results}
We perform multiple experiments to better understand different properties of our proposed approach. 
Namely, we study 
    \rev{(i) AC-MPC's performance on multiple drone racing tasks and its sample efficiency in comparison to AC-MLP policies, 
    (ii) the choice of matrix parameterization,
    (iii) its behaviour against unknown external disturbances,
    (iv) its robustness to changes in dynamic parameters,
    (v) the benefits of the Model-Predictive Value expansion,
    (vi) the interpretability properties of AC-MPC,
    (vii) its sensitivity to exploration hyperparameters,
    (viii) its real-world deployment.}
Each aspect corresponds to a different research question that we address in each of the following sections.
}
\subsection{Does our method improve performance and sample efficiency?}
\label{sec:horizontal_vertical}
We start with horizontal and vertical flight through gates.
The vertical task can show if the approach is able to find a solution that lies directly in the singularity of the input space of the platform since the platform can only generate thrust in its positive body Z direction.
When flying fast downwards, the fastest solution is to \extension{pitch and roll} the drone as soon as possible, direct the thrust downwards, and only then command positive thrust~\cite{foehn2021CPC}.
However, many approaches are prone to get stuck in a local optimum~\cite{romero2022mpcc}, where the commanded thrust is zero and the platform gets pulled only by gravity.
Fig. \ref{fig:horizontal_vertical_value_function} shows the simulation results of deploying the proposed approach, which was trained in the horizontal and vertical tracks (left and right side of Fig. \ref{fig:horizontal_vertical_value_function}, respectively).
We show velocity profiles and value-function profiles.
The value-function profiles have been computed by selecting a state of the platform in the trajectory and modifying only the position while keeping the rest of the states fixed.
For the horizontal track, we sweep only the XY positions, and for the vertical track, the XZ positions.
Additionally, 10 MPC predictions are shown and marked with Xs.
In these value function plots, areas with high values (in yellow) indicate regions with high expected returns.

In \extension{Fig. \ref{fig:horizontal_vertical_value_function} and in} the supplementary video, one can observe the evolution of the value function over time. 
Given the sparse nature of the reward terms (see Section \ref{sec:rewards}), one can observe that when a gate is successfully passed, the region of high rewards quickly shifts to guide the drone towards the next gate. 
This can be interpreted as a form of discrete mode switching enabled by the neural network cost map. 
Such mode-switching behavior is a challenging feat to accomplish using traditional MPC pipelines.
The intuition behind this is that the critic is able to learn long-term predictions, while the model-predictive controller focuses on the short-term ones, effectively incorporating two time scales.

\extension{
Additionally, we train our approach on a challenging track known as the SplitS track.
Designed by a professional drone racing pilot, this track is distinguished by its highly demanding SplitS maneuver, where the drone must navigate through two gates placed directly above one another successively.
This track was used in previous research as a benchmark, namely \cite{Song23Reaching, kaufmann23champion, foehn2021CPC, romero2022mpcc, mpcc_replanning}.

In Fig. \ref{fig:commands} we show the collective thrust commands of both approaches, AC-MLP vs AC-MPC. 
In this figure, one can notice how the AC-MPC can achieve saturation in a consistent way, while the AC-MLP approach also saturates but in a more irregular fashion.
This indicates that the MPC module within the AC-MPC framework, which includes explicit knowledge of both platform dynamics and its constraints, effectively and consistently utilizes the system's saturation limits. 
In contrast, the AC-MLP commands are generated directly by a neural network, resulting in behavior that is comparatively less consistent and less aligned to the platform’s true limits.

In Fig. \ref{fig:reward_evolution_tasks}, we can see the reward evolution for AC-MLP when compared with AC-MPC.
This comparison was conducted after optimizing the initial standard deviation hyperparameter for both approaches and choosing the best result for each.
Generally---as we also study in section \ref{sec:std_ablation}---AC-MPC performs better for lower exploration parameters, whereas AC-MLP needs higher exploration parameters.
Fig. \ref{fig:reward_evolution_tasks} shows that the asymptotic training performance and sample efficiency are improved when choosing AC-MPC, since it is able to leverage the prior information encoded in the dynamics within the differentiable MPC.
This is particularly evident in the reward evolution observed for the Vertical and SplitS tracks.
For simpler tasks, the prior knowledge introduced in AC-MPC is not strongly advantageous to performance.
For instance, in the Horizontal track, which is relatively simple, the training performance of AC-MLP is on par with AC-MPC, since the maneuvers that the policy needs to discover to perform the task do not require an elaborate exploration behavior.

In conclusion, the results presented in this section show that our method improves both performance and sample efficiency. 
By comparing AC-MPC against AC-MLP across various track complexities, and after hyperparameter tuning, we have demonstrated that AC-MPC can achieve superior performance and sample efficiency, especially in challenging tasks like the Vertical and SplitS tracks (see Fig. \ref{fig:reward_evolution_tasks}). 
This improvement is attributed to AC-MPC's ability to effectively leverage prior knowledge of the system dynamics through the embedded MPC module (see Fig. \ref{fig:commands}). 
These findings suggest that incorporating model-based control elements into reinforcement learning frameworks offers a promising direction for future research.
\begin{figure}[t]
    \centering
    \includegraphics[width=\linewidth]{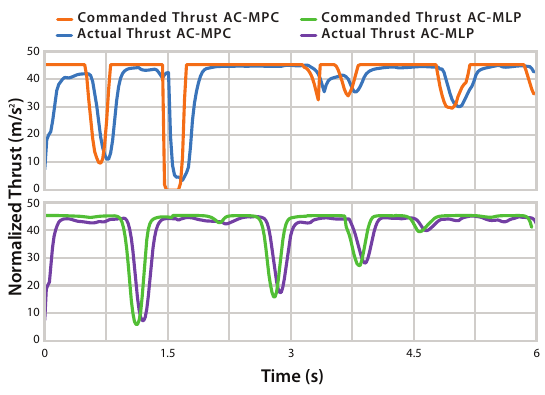}
    \caption{Evolution of the collective thrust command over time for both AC-MLP and AC-MPC during trials on the SplitS track. The figure demonstrates AC-MPC's ability to effectively utilize control input saturation due to its access to the system dynamics and control constraints. In contrast, AC-MLP exhibits less consistent saturation behavior.}
    \label{fig:commands}
\end{figure}
\begin{figure}[t]
    \centering
    \includegraphics[width=1.0\linewidth]{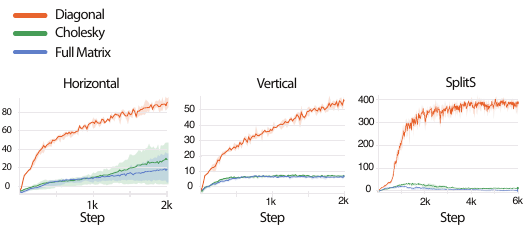}
    \caption{\rev{\revi{Reward evolution for agile quadrotor flight in the Split-S track}. Three different cost map representations are shown: Diagonal, Cholesky and Full Matrix.}}
    \label{fig:matrix_ablation}
\end{figure}
\begin{figure*}
    \centering
    \includegraphics[width=0.85\linewidth]{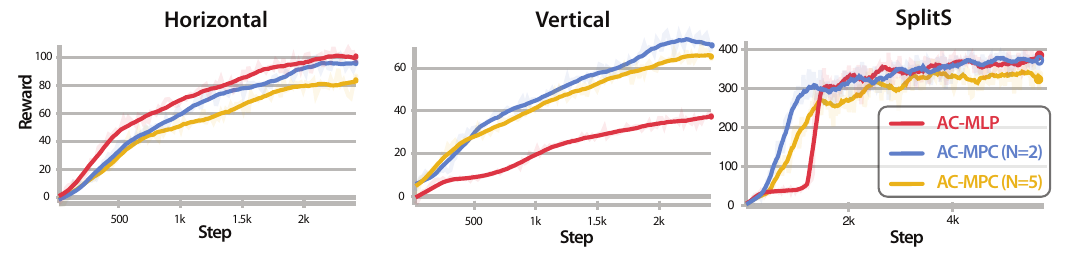}
    \caption{
    Reward evolution for \revi{drone racing in three} different tracks, for AC-MLP (for different N values of the horizon, $N = 2$ and $N = 5$) and AC-MPC. The values have been obtained after optimizing the initial exploration standard deviation for both approaches independently, and then selected the best result from each. For Vertical and SplitS tracks, one can observe how the AC-MPC approach is able to leverage its prior knowledge of the system and showcase improved learning. For the Horizontal track, because of its simplicity, the leverage of prior knowledge is diluted.
    \label{fig:reward_evolution_tasks}
    }
\end{figure*}
\begin{figure*}
    \centering
    \includegraphics[width=0.85\linewidth]{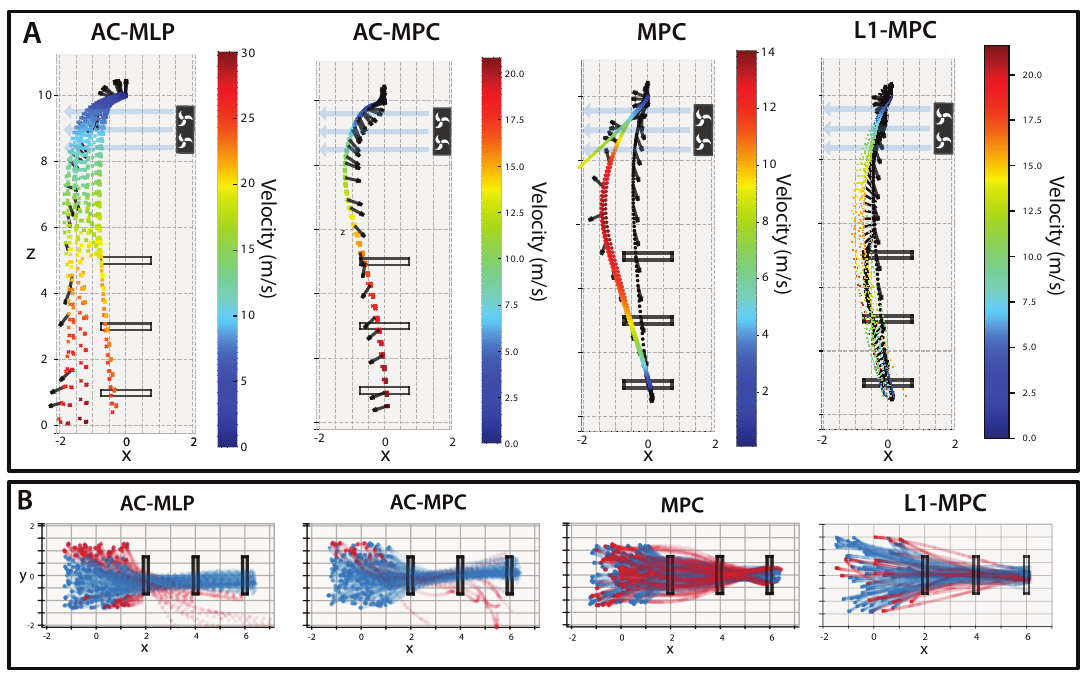}
    \caption{
    \rev{\textbf{Baseline comparisons between our AC-MPC, a standard PPO (termed AC-MLP), a standard tracking MPC and a L1-MPC, \revi{for the task of agile quadrotor flight.}}
    \textbf{(A)}: Robustness against wind disturbances (vertical track). All policies are trained without disturbances. Black arrows indicate the quadrotor's attitude. 
    \textbf{(B)}: Robustness against changes in initial conditions (horizontal track). Trajectories are color-coded, with crashed trajectories in red and successful in blue.}}
    \label{fig:ablation_study}
    \vspace{-20pt}
\end{figure*}
\begin{figure}[t]
    \centering
    \includegraphics[width=\linewidth]{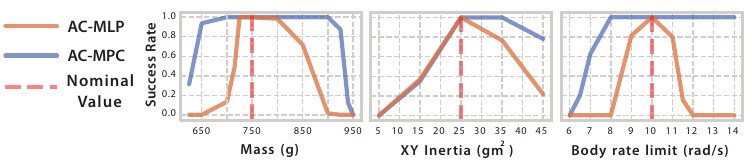}
    \caption{Success rate against variations of \revi{quadrotor's} mass, inertia, and body rate limit. Policies are trained in the SplitS track with a nominal value for these parameters. This nominal value is depicted in red. Then, these policies are deployed in 64 parallel environments, each environment having slightly different initial position. In the deployment stage, mass, inertia, and body rate limit are varied. The success rate is computed as the proportion of trajectories that don't have crashes. Because AC-MPC has the dynamics embedded in the network architecture, it shows to be more robust to these changes than its AC-MLP counterpart.
    \vspace{-16pt}}
    \label{fig:parameter_ablation}
\end{figure}
}
\rev{
\subsection{Does the choice of diagonal matrix affect performance?}

In this section, we investigate a critical design choice within our framework: the structure of the learnable cost matrix. 
Our goal is to empirically demonstrate that constraining this matrix to be diagonal does not negatively impact performance. 
On the contrary, we aim to show that this simplification makes the learning problem more tractable.

To this end, we compare our standard diagonal representation of the quadratic cost matrix Q against two more complex alternatives: a full matrix decomposition and a Cholesky decomposition. 
All three methods produce the positive semi-definite matrices required for the optimization problem, but they differ in the number of learnable parameters, which directly influences the complexity of the learning task.

For the Cholesky case, the free parameters are the non-zero parameters of a triangular matrix $A$, where the diagonal elements are positive and the rest are free to be positive or negative. In this case, the number of free parameters would be $N(N+1)/2$.
Similarly, for the full matrix case, the free parameters are all the elements of a matrix $A$, and therefore the number of free parameters would be $N^2$.
In both cases, the matrix $Q$ is parameterized as follows:
$$Q = A A^T,$$
This specific formulation is used to ensure that the resulting Q matrix is always positive semi-definite, a necessary condition for the stability of the underlying optimization problem.

Following this, we build three different cost map representations, mapping observations to cost function: \emph{Diagonal}, \emph{Cholesky} and \emph{Full Matrix}.
We train the three of them for three different tracks: Horizontal, Vertical and SplitS, and track the growth of the reward function to measure learning performance. 

As shown in the learning curves in Fig. \ref{fig:matrix_ablation}, the agent using the Diagonal representation is the only one capable of learning the required behavior, showing a clear and consistent increase in reward across all three tracks. 
In contrast, for both the Cholesky and Full Matrix cases, the network fails to learn within the same sample budget. 
We hypothesize that this outcome is a direct result of the higher learning difficulty imposed by the larger number of parameters in the more complex cost functions. 
This high-dimensional search space likely prevents the learning algorithm from finding a useful gradient, demonstrating that the constrained and more manageable problem offered by the diagonal matrix helps the training.
}
\subsection{Is our approach robust to external disturbances?}
\rev{We perform various studies where the standard actor-critic PPO architecture~\cite{Song23Reaching} (labeled as \emph{AC-MLP}), a standard tracking MPC and a state-of-the-art adaptive L1-MPC~\cite{l1mpc} controller are compared to our approach (labeled as \emph{AC-MPC}) in terms of generalization and robustness to disturbances.}
Both the AC-MLP and AC-MPC approaches are trained with the same conditions (reward, environment, observation, simulation, etc.).
\rev{
It is important to note that since we aim to test for out-of-distribution behaviour, the disturbances are not present at training time, only during testing.
}
All these evaluations are conducted using the high-fidelity BEM simulator~\cite{bem}.
\rev{Both the MPC and the L1-MPC approaches track a time-optimal trajectory obtained from~\cite{foehn2021CPC}.}

\begin{scriptsize}
\begin{table}
    \caption{\rev{Comparison: Success Rate and average velocity for agile flight through the Horizontal, Vertical and Vertical with wind tracks.}}
    \label{tab:performance}
    \scriptsize %
    \begin{tabularx}{\linewidth}{l|S[table-format=2.2]S[table-format=2.2]|S[table-format=2.2]S[table-format=2.2]|S[table-format=2.2]S[table-format=2.2]}
        \toprule
        & \multicolumn{2}{c}{\textbf{Horizontal}} & \multicolumn{2}{c}{\textbf{Vertical}} & \multicolumn{2}{c}{\textbf{Vertical Wind}} \\
        \cmidrule(lr){2-3} \cmidrule(lr){4-5} \cmidrule(lr){6-7}
        & {\textbf{SR} [\%]} & {$\vec{v}$ [m/s]} & {\textbf{SR} [\%]} & {$\vec{v}$ [m/s]} & {\textbf{SR} [\%]} & {$\vec{v}$ [m/s]} \\
        \midrule
        \textbf{AC-MLP} & 74.78 & 7.74 & 53.61 & 10.56 & 6.5 & 10.67 \\
        \textbf{MPC} & 64.94 & 4.15 & 72.27 & 4.25 & 0.0 & 6.44 \\
        \textbf{L1-MPC} & 80.0 & 5.24 & \textbf{100.0} & 4.61 & 41.66 & 8.69 \\
        \textbf{AC-MPC} & \textbf{90.37} & 6.51 & 64.47 & 10.05 & \textbf{83.33} & 10.76 \\
        \bottomrule
    \end{tabularx}
\end{table}
\end{scriptsize}

In terms of disturbance rejection and out-of-distribution behavior, we conduct three ablations, shown in Fig. \ref{fig:ablation_study} and Table \ref{tab:performance}).
In Fig. \ref{fig:ablation_study}A (and the \emph{Vertical Wind} column of Table \ref{tab:performance}), we simulate a strong wind gust that applies a constant external force of 11.5 N (equivalent to 1.5x the weight of the platform).
This force is applied from $z=10m$ to $z=8m$.
We can see how neither the AC-MLP nor the MPC policies can recover from the disturbance and complete the track successfully.
\rev{
The adaptive L1-MPC controller is able to overcome the disturbance and pass through the gates with a success rate of 41.66\% and an average speed of 8.69 m/s.

On the other hand, AC-MPC achieves a higher success rate and speeds (83.33\% and 10.76 m/s, as shown in Table \ref{tab:performance}), and exhibits more consistency among repetitions.
This showcases that incorporating an MPC block enables the system to achieve better out of distribution behaviour.
}

For the \emph{Vertical} and \emph{Horizontal} experiments in Table \ref{tab:performance}, we simulate 10000 iterations for each controller where the starting points are uniformly sampled in a cube of 3m of side length where the nominal starting point is in the center.
In the \emph{Horizontal} case, the results are shown in Fig. \ref{fig:ablation_study}B.
It is important to highlight that during training of AC-MLP and AC-MPC, the initial position was only randomized in a cube of 1m of side length.
The successful trajectories are shown in blue in Fig. \ref{fig:ablation_study}B, while the crashed ones are shown in red. In Table \ref{tab:performance}, we can observe that the AC-MPC presents a higher success rate than AC-MLP in both experiments.
\rev{
One can also see that AC-MPC has a higher success rate than both the MPC and the L1-MPC approaches in the \emph{Horizontal task}, but this is not the case in the \emph{Vertical} task.
}
\rev{
The reason behind this is that in the \emph{Vertical} task, the MPC and L1-MPC are not able to track the solution that turns the drone upside down, therefore resulting in the sub-optimal solution of setting all thrusts to near-zero state and dropping only by the effect of gravity, which results in slower but safer behavior. This is evident by looking at the average speed column in Table \ref{tab:performance}.
}

\rev{
It is important to clarify that the tracking MPC approaches have difficulties in this task because the trajectories to be tracked are time-optimal trajectories.
This means that they command the platform to go from hover to 100\% thrust and body rates commands directly from the start.
This makes these trajectories extremely difficult or impossible to track in the presence of any disturbance or model mismatch.
}

These experiments provide empirical evidence showing that AC-MPC exhibits enhanced performance in handling unforeseen scenarios and facing unknown disturbances, which makes it less brittle and more robust.

\extension{
\subsection{Is our approach robust to changes in the dynamics?}
\rev{Our AC-MPC approach integrates a model-based controller that explicitly accounts for system dynamics, and therefore it benefits from the possibility of changing dynamic parameters without retraining, up to certain extent.
The question arises of whether AC-MPC is less sensitive to dynamic parameters changes when compared to AC-MLP.}
To investigate this, we conduct a series of experiments in which key dynamic parameters are systematically varied. 
Both policies are trained on the SplitS track using nominal parameter values, after which they are evaluated across a range of perturbed parameters for both AC-MLP and AC-MPC.
To quantify the robustness of these policies, we deploy them from 64 different positions, equally spread on a cube of $0.5\, m$ of side length.
Success is defined as the percentage of trajectories that successfully navigate through all gates of the SplitS track. Fig.~\ref{fig:parameter_ablation} presents the success rate as a function of variations in mass (up to $+27$\%), in XY inertia (up to $\pm 80$\%) , and body rate limits (up to $\pm 40\%$).

The results demonstrate that the AC-MPC exhibits superior generalization to dynamic parameter variations compared to the AC-MLP.
This improved robustness is attributed to the incorporation of a dynamic model within the differentiable MPC block, which allows for adjustments in response to perturbed dynamics during deployment without retraining.
}
\subsection{What do we gain from Model-Predictive Value Expansion?}
  \label{sec:mpve_ablation}
  In this section, we conduct empirical experiments to evaluate the effect of the AC-MPC extended with Model-Predictive Value Expansion algorithm (introduced in Section~\ref{sec:mpve}) in terms of training performance.
  To isolate the effect of MPVE and enable clear reward evaluation, we employ a straightforward hovering task for the drone. Here, the drone starts at a random position and orientation within a 1x1x1 meter cube around a designated equilibrium point (0, 0, 5 meters). The objective is to stabilize the drone at this point with zero velocity and a perfectly upright orientation (hover).
  \rev{We compare the training performance of AC-MPC-MPVE against AC-MLP and AC-MPC. All algorithms are trained with a fixed sample budget of 50 steps, corresponding to roughly \revi{1.25} million samples collected from the environment.}
  \revi{For these experiments, the horizon of the differentiable MPC is set to $N=3$, and the MPVE $H$ is set to the same value.}
  Because the MPVE algorithm primarily affects the learning of the critic, which uses the Generalized Advantage Estimation (GAE) algorithm~\cite{schulman2015gae}, we explore the impact of our extension for various values of the temporal difference parameter $\lambda$ in TD($\lambda$).
  \rev{The parameter $\lambda$ determines the trade-off between using the Bellman error and using the Monte-Carlo estimates for the value function approximation.}
  A lower value of $\lambda$ favors lower variance but higher bias by relying more on immediate rewards, while higher values reduce bias but increase variance by incorporating more long-term information.
  To investigate this, we experiment with six different values for $\lambda$: 0.0, 0.2, 0.4, 0.6, 0.8, and 1.0. For each value, we repeat the training process three times with different random seeds.

\begin{figure}[t]
    \centering
    \includegraphics[width=\linewidth]{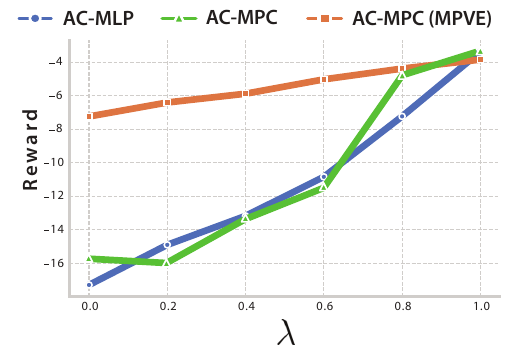}
    \caption{\rev{\revi{For the quadrotor stabilization task, we show the} reward at 50 steps of training for AC-MLP, AC-MPC and AC-MPC~(MPVE), depending on the $\lambda$ used in the TD($\lambda$). The policies are trained with a sample budget of 50 steps and the final reward is recorded, for different values of $\lambda$, for both AC-MPC~(MPVE) and AC-MLP.}}
    \label{fig:rew_lambda}
  \end{figure}

  Our results (visualized in Fig.~\ref{fig:rew_lambda}) demonstrate that the sample efficiency gains achieved by MPVE are more significant for lower lambda values. This can be explained by the way the value function leverages data from the rollout buffer. Higher lambda values incorporate data from a larger portion of the rollout, increasing the influence of Monte Carlo estimates and diluting the relative impact of the information provided by MPVE predictions.
  In contrast, lower lambda values lead to stronger benefits from MPVE. This suggests that our algorithm is particularly advantageous for scenarios where the value function is trained with TD(0). This approach is often preferred when memory limitations exist (as it avoids storing the entire rollout buffer) or for non-episodic tasks.

\begin{figure}[t]
    \centering
    \includegraphics[width=\linewidth]{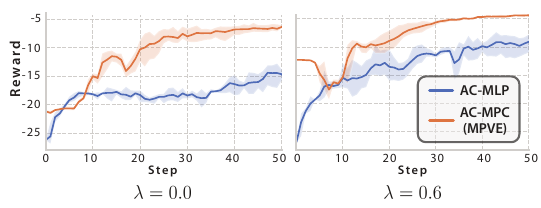}
    \caption{Reward evolution for \revi{the quadrotor stabilization task, for} the cases of $\lambda = 0.0$ and $\lambda = 0.6$, for AC-MLP and for AC-MPC~(MPVE).}
    \label{fig:rew_lambda_evolution}
    \vspace{-28pt}
  \end{figure}

  In Fig.~\ref{fig:rew_lambda_evolution}, we showcase the evolution of the episode returns for AC-MLP and AC-MPC-MPVE under two different values of $\lambda$ and for the same task of drone stabilization.
  In both settings, our approach substantially outperforms the baseline in terms of asymptotic performance and sample efficiency, highlighting the benefits of model-predictive value expansion.

  \rev{
  We would like to note that while we modify the value of $\lambda$ in this section for the sake of exposing the influence of MPVE, the value of $\lambda$ in the rest of the article is kept constant and equal to $\lambda=0.95$. 
  This is a very common choice of $\lambda$ in general TD($\lambda$) architectures and empirically gives the most stable learning when roll-outs are long and the task demands a large planning horizon, as is the case in drone racing.
  As one can see in Fig. \ref{fig:rew_lambda}, as $\lambda$ approaches one, the returns of all three controllers become indistinguishable: with an almost Monte-Carlo target, the critic already receives a near-complete estimate of the tail cost, so the extra predictions injected by MPVE provide little additional benefit.
  Table \ref{tab:ppo_params} shows an overview of the hyperparameters used for training both AC-MPC and AC-MLP for the rest of the article.
  }

\begin{table}[ht]
    \centering
    \setlength{\tabcolsep}{2pt} %
    \caption{AC-MLP and AC-MPC training hyper-parameters}
    \begin{tabular}{@{} l r @{}} %
        \toprule
        \textbf{Setting}                     & \textbf{Value} \\
        \midrule
        Discount factor $\gamma$             & 0.98 \\ %
        GAE $\lambda$                        & 0.95 \\
        Steps per update                     & 250 \\
        Mini-batch size                      & 25\,000 \\
        SGD epochs                           & 10 \\
        Clip range                           & 0.2 \\
        Learning rate                        & $3{\times}10^{-4}\!\rightarrow\!10^{-5}$ \\
        Entropy coef.                        & 0.001 \\
        Value-loss coef.                     & 0.5 \\
        Grad. clip                           & 0.5 \\
        Init.\ log-std (AC-MPC / AC-MLP)     & $-1.2$ / $-0.5$ \\
        Policy net size                      & {[}512, 512{]} \\
        Value net size                       & {[}512, 512{]} \\
        \bottomrule
    \end{tabular}
    \label{tab:ppo_params}
\end{table}

\rev{Additionally, in table \ref{tab:comp_times} we added the training time difference between adding MPVE and not adding it. 
It is important to note that since the actor network is not modified, the inference time remains identical for both algorithms.
}
\begin{table}[h]
  \centering
  \caption{\rev{Training time of the two algorithms for 2M steps.}}
  \begin{tabular}{@{}l c@{}}
    \toprule
    \textbf{Method} & \textbf{Time} \\
    \midrule
    AC-MPC-MPVE & 1\,h 19\,min \\
    AC-MPC      & 0\,h 50\,min \\
    \bottomrule
  \end{tabular}
  \label{tab:comp_times}
\end{table}
  

\extension{
\subsection{Can we interpret the internal workings of AC-MPC?}
\label{sec:hessian_cost}
This section investigates a potential connection between the critic network's output, $V(\vec{x})$, and the learned cost matrix $Q(\vec{x})$. To simplify the analysis, we introduce three key assumptions. 
First, we assume that the observations $s_k$ are equivalent to the state of the system $x_k$.
Second, the reward function is assumed to be quadratic in the state, expressed as $r(\vec{x}) = -\vec{x}^T \vec{Q} \vec{x}$. This formulation captures the objective of stabilizing the system around a hover point. 
And third, we restrict the Model Predictive Control (MPC) horizon length to $T = 2$.
\rev{
These assumptions are chosen to create a simplified, tractable setting that mirrors the structure of a classic Linear Quadratic Regulator (LQR) problem, where the relationship between the value function and the quadratic cost is well-understood. 
While our system remains non-linear, this setup allows us to isolate and probe the fundamental connection between the critic’s learned value function and the actor's learned MPC cost.
}

Given the focus on drone dynamics and a stabilization task, we expect the system to primarily operate near hover states. This implies that the learned value function should also be approximately quadratic.
The central question we aim to address here is whether a relationship exists between the Hessian (the matrix containing all second-order partial derivatives) of this value function and the learned cost matrix, $Q$.
The general infinite horizon problem for an MPC can be written as:
\begin{align*}
  \pi^*(x_0) &= \argmin_u \sum_{k=0}^\infty l(x_k,u_k)\\
  &= \argmin_u \sum_{k=0}^{N-1} l(x_k,u_k) + \sum_{k=N}^{\infty} l(x_k,u_k)\\
  &= \argmin_u \sum_{k=0}^{N-1} l(x_k,u_k) - V^*(x_N) \\
  &= \argmin_u \sum_{k=0}^{N-1} l(\tau_k) - V^*(f(\tau_{N-1})),
\end{align*}
where  $\tau_{k} = \mat{x_{k}$, $u_{k}}$ is a state-action pair at time $k$ and the value function term is negative because we have defined it in the context of RL, where it is maximized.

\begin{figure*}
    \centering
    \includegraphics[width=0.9\linewidth]{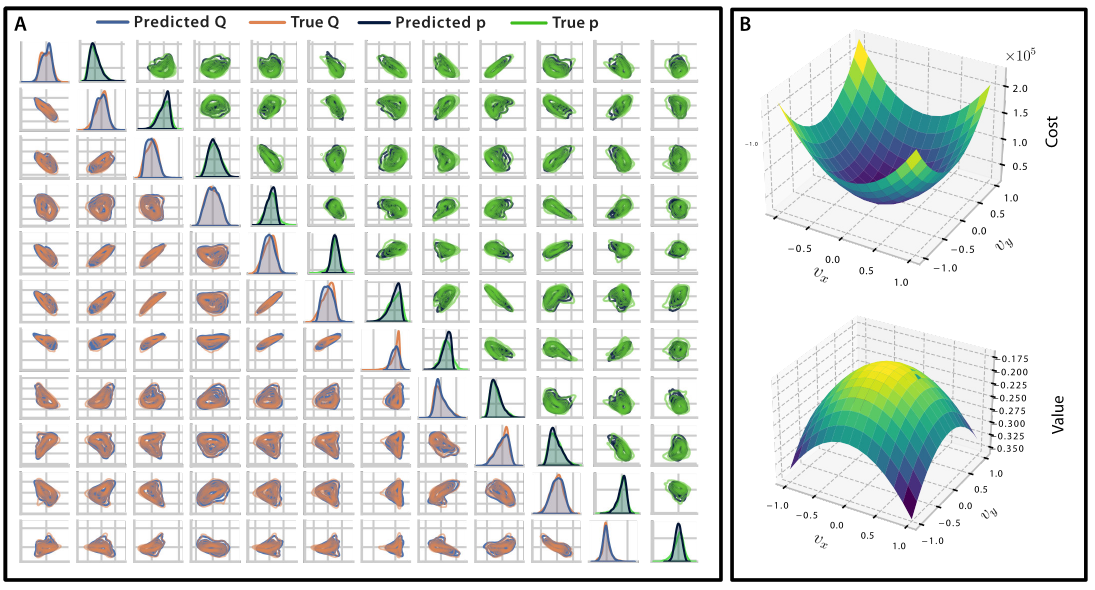}
    \caption{
    \textbf{Empirical probing of the correlation between the Hessian and the gradient of the learned critic $V(x)$ and the learned differentiable MPC cost terms $Q_N$ and $p_N$, \revi{for the task of quadrotor stabilization}.}
    \textbf{(A)}: We train a linear classifier to infer $Q_N$ and $p_N$ from the Hessian of $V(x)$ and show the prediction from different 2D projections of $Q_N$ and $p_N$. We show how the ground truth can be predicted with high accuracy by only using a small linear classifier, which indicates that the data learned by the critic and $Q_N$ and $p_N$ are highly correlated.
    \textbf{(B)}: Top: 3D plot of $\tau_N^T Q_N \tau_N + \tau_N^T p_N$ for the state dimensions of $v_x$ and $v_y$ after training. Bottom: 3D plot of $V(x)$ for the same dimensions. Apart from the change in sign and in scale, the similarity between both shapes indicates a high correlation between the value function and the learned cost parameters.}
    \label{fig:probing_hessian}
    \vspace{-20pt}
\end{figure*}

\rev{
In this formulation, the term $V^*(x_N)$  represents the optimal cost-to-go from the end of the finite horizon N. 
The hypothesis of our analysis is that the critic, through reinforcement learning, learns an approximation of this infinite-horizon value function. 
The actor, in turn, learns the quadratic cost parameters ($Q_N, p_N$) that best represent the local, second-order approximation of this value function at the horizon's boundary.
}

On the other hand, the quadratized problem that we solve in the differentiable MPC block has the following form:
\begin{align}
     \sum_{k = 0}^{N} \mat{x_k \\ u_k}^T Q_k \mat{x_k  \\ u_k} + p_k \mat{x_k \\ u_k}.
\end{align}
This motivates the investigation of a potential relationship between the value function $V(x)$ learned by the critic and the quadratic ($Q_N$) and linear ($p_N$) terms learned by the actor within the differentiable MPC framework. 
In particular, we aim to identify a connection between the first and second derivatives of the value function and $p_N$ and $Q_N$, respectively. To achieve this, we can introduce the second-order Taylor Series approximation of the value function around the state-action pair $\tau_{N-1}$:
\begin{align*}
V(f(\tau_{N-1})) \bigg\rvert_{\tau_{N-1} = \vec{d}} &\approx V(f(\vec{d})) \\
&\quad + \tau_{N-1}^T \frac{\partial^2 V(f(\tau_{N-1} = \vec{d}))}{\partial \tau_{N-1}^2} \tau_{N-1} \\
&\quad + \frac{\partial V(f(\tau_{N-1} = \vec{d}))}{\partial \tau_{N-1}} \tau_{N-1} \\
&= V(f(\vec{d})) + \tau_{N-1}^T H_V \tau_{N-1} + \Delta_V \tau_{N-1},
\end{align*}
where $f(\cdot)$ represents the system dynamics function, $\vec{d}$ denotes a specific state-action pair, $H_V$ is the Hessian of the value function evaluated at $\vec{d}$, $\Delta_V$ represents the first-order term of the Taylor expansion.
Because we need to compute the gradient and the Hessian of the value function composed with the dynamics, here are the expressions for $\Delta_V$ and $H_V$.
\begin{align}
\label{eq:delta_v}
\Delta_V &= \frac{\partial}{\partial \tau_{N-1}} (V(f(\tau_{N-1}))) \nonumber
= \frac{\partial}{\partial \tau_{N-1}} (V(x_N))  \\ \nonumber
&= \frac{\partial V(x_N)}{\partial x_N} \frac{\partial x_N}{\partial \tau_{N-1}}
= \frac{\partial V(x_N)}{\partial x_N} \frac{\partial f(\tau_{N-1})}{\partial \tau_{N-1}} \\
&= \frac{\partial V(x_N)}{\partial x_N} \mat{A_{N-1}, B_{N-1}},
\end{align}
where $\mat{A_{N-1}, B_{N-1}}$ are the matrices associated to the linearized dynamics at timestep $N-1$, and $x_N = f(\tau_{N-1})$, $\frac{\partial V(x_N)}{\partial x_N}$ is the gradient of the value function with respect to its inputs. For $H_V$:

\begin{align}
\label{eq:h_v}
H_V &= \nonumber
\frac{\partial}{\partial \tau_{N-1}} (\Delta_V) \\ \nonumber
&= \frac{\partial}{\partial \tau_{N-1}} \left( \frac{\partial V(x_N)}{\partial x_N} \right) \frac{\partial f(\tau_{N-1})}{\partial \tau_{N-1}} \\ \nonumber
&+ \frac{\partial V(x_N)}{\partial x_N} \cancelto{0}{\frac{\partial^2 f(\tau_{N-1})}{\partial \tau_{N-1}^2}} \\ \nonumber
&= \frac{\partial x_N^T}{\partial \tau_{N-1}} \frac{\partial^2 V(x_N)}{\partial x_N^2} \frac{\partial x_N}{\partial \tau_{N-1}} \\ \nonumber
&= \frac{\partial f(\tau_{N-1})^T}{\partial \tau_{N-1}} \frac{\partial^2 V(x_N)}{\partial x_N^2} \frac{\partial f(\tau_{N-1})}{\partial \tau_{N-1}} \\
&= \mat{A_{N-1}\\B_{N-1}}\frac{\partial^2 V(x_N)}{\partial x_N^2}  \mat{A_{N-1}, B_{N-1}},
\end{align}

where we have assumed that the second derivative of the dynamics is zero, and $\frac{\partial^2 V(x_N)}{\partial x_N^2}$ is the Hessian of the value function with respect to its inputs.

In order to empirically show the relationship between the Hessian of the value function and what is learned in $Q_N$ and $p_N$, we perform a similarity study using linear probes.
Particularly, we first train an AC-MPC agent to perform the stabilization at hover task, introduced in Section \ref{sec:mpve_ablation}.
We choose this task here because, since the reward function is quadratic, it results in an optimization landscape that is easier to interpret.
Since we need the derivative of the value function with respect to the full state, the critic needs to be trained with full state information.
After training, we collect a dataset of uniformly distributed $2^{19} = 524288$ datapoints, where every datapoint corresponds to a sample of $Q_N$, $p_N$, $H_V$ and $\Delta_V$, as per Eq. \eqref{eq:h_v} and \eqref{eq:delta_v}.
Then, with this dataset, we train a single linear layer (without non-linearity) to predict the elements of the learned $Q_N$ and $p_N$ matrices from the elements of the $H_V$ and $p_V$, respectively.
In the hypothesized case where the information contained in the learned cost is related to the information encoded by the gradient and Hessian of the value function, we expect the regression to have a high accuracy.

In Fig. \ref{fig:probing_hessian}A, we show the true and predicted values of $Q_N$ and $p_N$. 
In order to show that the data fully correlates in all dimensions, the data is presented in a lower triangular matrix for $Q$ and in an upper triangular matrix for $p$.
Every row and column of this matrix is composed of the elements of the state space $x$, therefore showing the correlation from different 1-dimensional cuts -- for the diagonal elements -- and 2-dimensional cuts -- for the rest.
One can therefore see a high correlation between the predicted and the true values, therefore confirming a strong relationship between the value function and the learned cost parameters.
An extra experiment was conducted where we have tried to infer the same Q and p values, but from a random dataset, as a validation that there is no overfitting behavior happening. Indeed, in this case the linear classifier was not able to predict the true labels at all.

In conclusion, this section directly investigates and reveals a relationship within the internal workings of AC-MPC.
By demonstrating a strong correlation between the Hessian and gradient of the learned value function and the learned cost parameters $Q_N$ and $p_N$, we have shown that the critic effectively learns information related to the curvature of the cost landscape used by the MPC module. 
This provides valuable insight into how the critic informs the MPC controller and suggests that the learned value function encodes information about the system's local dynamics and cost structure.
}

\extension{
\subsection{How sensitive is our approach to exploration hyperparameters?}
\label{sec:std_ablation}
Since our approach incorporates a model of the dynamics within the optimization problem, our policy has access to prior information before the training phase, and hence leverages prior knowledge of the dynamics in its architecture.
\rev{In contrast, its AC-MLP counterpart consists of a randomly initialized MLP policy, which has no prior built-in knowledge about the system.}
This raises questions about the extent of exploration needed to discover an optimal policy and how the exploration parameter can be minimized to achieve optimal performance.
To explore this, we study the effect of the initial policy standard deviation hyperparameter on the performance of both AC-MLP and AC-MPC.
\rev{
This parameter controls how much Gaussian noise is injected at the control input level of the policy.
This means that for the AC-MPC, during training the noise is injected on top of the solution that comes from the optimizer.
Therefore, excessive noise can result in these solver solutions being almost entirely disregarded.
}
In Fig.~\ref{fig:std_ablation}, we present reward plots for the training on the Vertical track under different values of this hyperparameter.
The results show that AC-MLP requires a higher degree of exploration to identify the optimal solution. 
This indicates that AC-MPC effectively utilizes the prior dynamics knowledge, resulting in improved performance with reduced exploration.
More interestingly, Fig.~\ref{fig:std_ablation} shows that AC-MPC is less sensitive to the choice of this hyperparameter and yields high performance despite very low exploration values.
On the other hand, AC-MLP fails to reach high-reward regions when trained with low values and generally underperforms in comparison to the proposed method.
\rev{
As a conclusion of this study, the main guideline to tune the proposed AC-MPC is to use less exploration noise than generally needed for AC-MLP.
Table \ref{tab:ppo_params} shows the hyperparameters used for both AC-MPC and AC-MLP, where the main difference is the initial exploration hyperparameter.
}
\begin{figure}[ht]
    \centering
    \includegraphics[width=\linewidth]{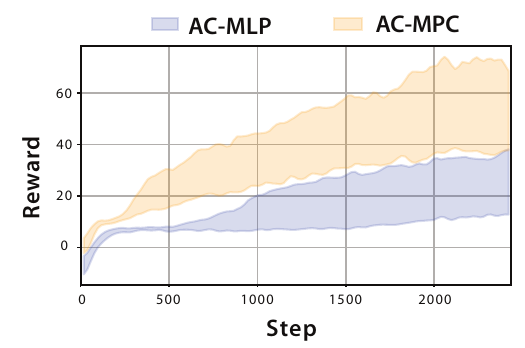}
    \caption{Comparison of reward values during training for different exploration parameters [0.22, 0.3, 0.44, 0.6] for both AC-MLP and AC-MPC, for the Vertical track.}
    \label{fig:std_ablation}
\end{figure}
}

\subsection{How does our method perform when deployed in the real-world?}
\label{sec:real_world}
We test our approach in the real world with a high-performance racing drone. 
\extension{In order to test simulation to real world transfer,} we deploy the policy with two different race tracks: Circle track and SplitS track.
We use the Agilicious control stack~\cite{agilicious} for the deployment. 
The main physical parameters and components of this platform are based on~\cite{Song23Reaching}, under the name \emph{4s drone}.
\begin{figure}
    \centering
    \includegraphics[width=0.95\linewidth]{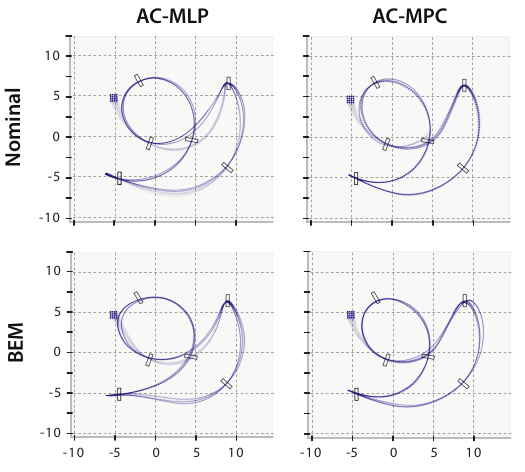}
    \caption{Top view of the performance of AC-MLP and AC-MPC policies deployed in both nominal and a realistic simulator for \revi{quadrotor racing in} the SplitS track. The policies are tested for 64 different initial positions, distributed in a cube of $0.5 m$ of side length.}
    \label{fig:simulation_splits}
    \vspace{-20pt}
\end{figure}
\begin{figure}
    \centering
    \includegraphics[width=0.95\linewidth]{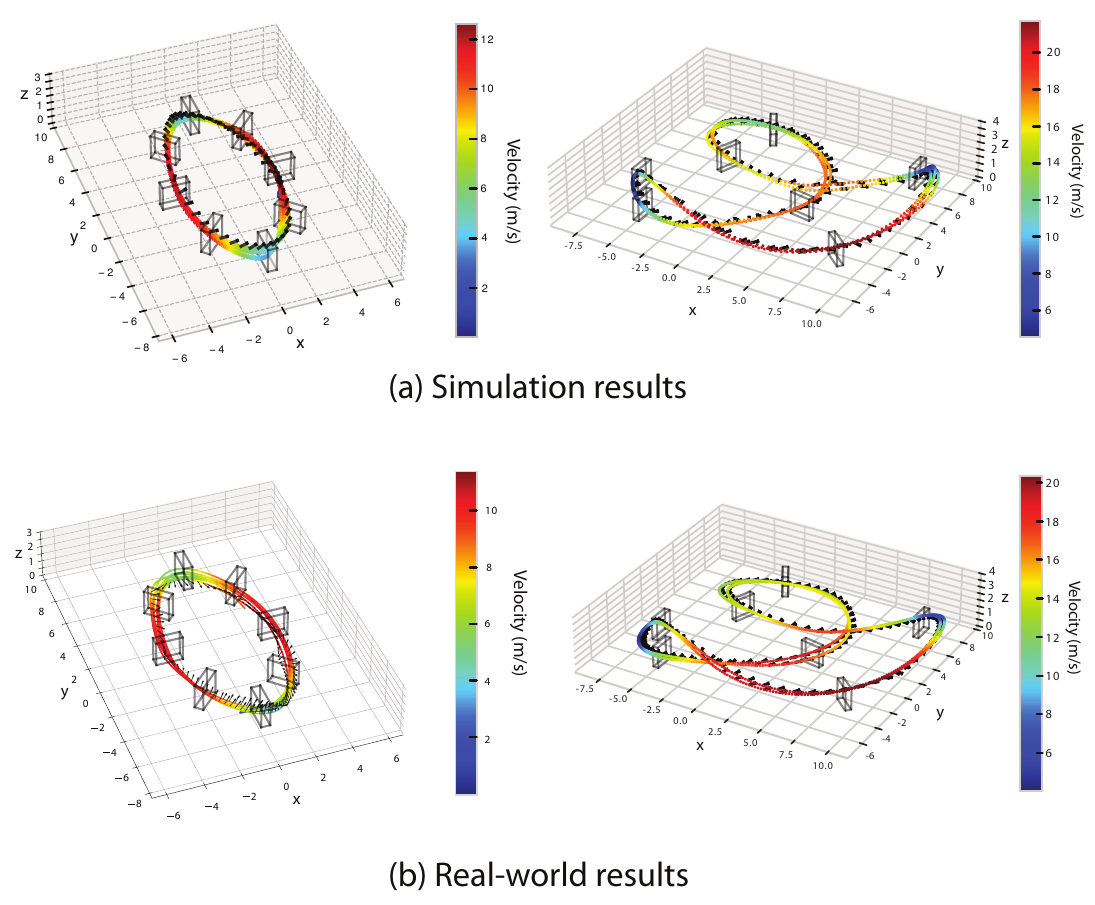}
    \caption{AC-MPC trained for the task of agile flight in complex environments. On the left is the Circle track, and on the right is the SplitS track for both the real world and simulation. These figures show how our approach is able to be deployed in the real world and how they transfer zero shot from simulation to reality. The plots show the flown trajectories by our quadrotor platform, recorded by a motion capture system.}
    \label{fig:sim2real_transfer}
\end{figure}

\extension{
Additionally, we perform a lap time comparison in the SplitS track.
We train both AC-MPC and AC-MLP in the task of drone racing and execute different runs in two different simulation fidelities: nominal dynamics and realistic simulator.
This realistic simulator, called NeuroBEM, and based in blade element momentum theory, comes from \cite{bem}.
\rev{
We would like to note that quadcopters are complex robots and many difficulties arise when transferring policies from simulation to the real hardware, especially given the inherent aggressiveness of time-optimal policies.
To address these limitations, we follow the sim2real techniques used in \cite{kaufmann23champion}, where we first train policies in a non-fine-tuned simulator, then deploy these policies and collect more data.
This data is then used to refine and develop a higher fidelity simulator, which is then used to train the final policy. 
On top of this, we use a significant amount of domain randomization in key parameters such as thrust map, mass, latency, drag and initial state.
Finally, a term in the reward function that smoothes the control actions is needed to avoid jitter of the platform and ensure a safe deployment.
}
In Fig. \ref{fig:simulation_splits} we show a top view of the deployment of both AC-MLP and AC-MPC in both of the aforementioned simulators.
The policies are tested for 64 different initial positions, contained in a cube of $0.5$ m of side length.
Additionally, Fig. \ref{fig:sim2real_transfer} illustrates trajectories flown in two different tracks (Circle track and SplitS track) in simulation and the real world. 
Not only can our policy transfer to the real world without any fine-tuning, but it can do so while at the limit of handling, achieving speeds up to 21 m/s.

In Table \ref{tab:laptimes}, we show the results of this comparison in terms of lap time and success rate, including both simulation modalities and real-world results.
\begin{table}
\centering
\caption{Performance comparison between AC-MLP and AC-MPC in the SplitS track}
\label{tab:laptimes}
\resizebox{0.5\textwidth}{!}{%
\begin{tabular}{c c c | c c}
    \toprule
    & \multicolumn{2}{c|}{AC-MLP} & \multicolumn{2}{c}{AC-MPC (N=2)} \\
    \cmidrule(r){2-3} \cmidrule(l){4-5}
    \textbf{Drone Model} & Lap Time [s] & Success Rate [\%] & Lap Time [s] & Success Rate [\%] \\ 
    \midrule
    Nominal & 5.09 $\pm$ 0.008 & 100.0 & 5.13 $\pm$ 0.008 & 100.0 \\ 
    Realistic & 5.179 $\pm$ 0.01 & 100.0 & 5.24 $\pm$ 0.01 & 100.0 \\ 
    Real World & 5.39 $\pm$ 0.08 & 85.7 & 5.4 $\pm$ 0.082 & 87.5 \\ 
    \bottomrule
\end{tabular}%
}
\end{table}
As observed in the lap time results, both AC-MPC and AC-MLP exhibit similar performance in terms of lap time, indicating that they are largely on par in this aspect. 
This similarity is expected, given that their asymptotic performance and final reward values are quite close, as illustrated in Figure \ref{fig:reward_evolution_tasks}.
However, it is important to note that our approach not only achieves on-par performance but also demonstrates additional \rev{properties in out-of-distribution scenarios} (see Fig. \ref{fig:ablation_study}, in Horizontal and Vertical tracks), and changes in the system dynamics (see Fig. \ref{fig:parameter_ablation}, in SplitS track). 
This indicates that our method maintains its effectiveness under a wider range of conditions and uncertainties, while still matching the performance of the neural network-based architecture.
}
\subsection{How do training and inference times compare to standard actor-critic architectures?}
\label{sec:times}
In Table \ref{tab:times}, we show the training times (SplitS track) and the forward pass times for AC-MLP and for the proposed AC-MPC for different horizon lengths.
\extension{
As can be seen, while AC-MPC offers performance advantages, it incurs a higher computational cost. 
Training time is increased by a factor of approximately 30 (for a horizon length of N=2) due to the batched optimization required for each forward and backward pass. 
This computational overhead also affects online performance, resulting in a slower forward pass. 
Although the current Python implementation\footnote{\url{https://locuslab.github.io/mpc.pytorch/}} of the differentiable MPC algorithm \cite{amos2018differentiable} achieves a sufficient deployment frequency of 50Hz for our system, optimized C/C++ implementations would significantly reduce training time and enable higher frequency deployments. 
Note, however, that this computational overhead did not affect closed-loop performance, as the 13.5ms execution time remained well within our 20ms control time step budget necessary to deploy the system successfully in the real world.
\rev{
Furthermore, although classical MPC often requires horizons longer than $N=2$, our architecture uses a trained critic network to learn and provide the long-term reasoning to the short-horizon solver. 
Our analysis in Section \ref{sec:hessian_cost} shows that the learned MPC cost matrices are highly correlated with the Hessian of the critic's value function. 
This indicates that the critic learns an accurate, infinite-horizon cost-to-go, which it then uses to inform the differentiable MPC.
}

}

\begin{scriptsize}
\begin{table}[htbp]
  \centering
  \caption{Solve times and inference times for different variations.}
  \begin{tabular}{lcc}
    \toprule
    & Training time & Inference time \\
    \midrule
    AC-MLP & 21m & 0.5 $\pm$ 0.037 ms \\
    AC-MPC (N=2) & 11h:30m & 13.5 $\pm$ 1.1 ms \\
    AC-MPC (N=5) & 22h:6m & 37.5 $\pm$ 14.5 ms \\
    AC-MPC (N=10) & 39h:36m & 69.9 $\pm$ 22 ms \\
    \final{AC-MPC (N=50)} & \final{-} & \final{210.32 $\pm$ 22.4 ms} \\
    \bottomrule
  \end{tabular}
  \label{tab:times}  
\end{table}
\end{scriptsize}

\section{Discussion and Conclusion}
\label{sec: conclusion}
This work presented a new learning-based control framework that combines the advantages of differentiable model predictive control with actor-critic training. 
\extension{
Furthermore, we showed that \revi{for quadrotor racing tasks,} AC-MPC can leverage the prior knowledge embedded in the system dynamics to achieve better training performance \revi{when compared to the main PPO baseline (AC-MLP)}, to better cope with out-of-distribution scenarios, \rev{and its ability to deal with variations in the nominal dynamics, without any further re-training}.
Through empirical analysis, the interpretable nature of the differentiable MPC embedded in the actor-critic architecture allows us to reveal a relationship between the learned value function and the learned cost functions. This provides a deeper understanding of the interplay between the reinforcement learning (RL) and MPC, offering valuable insights for future research.
Additionally, our approach achieved zero-shot sim-to-real transfer, demonstrated by successfully controlling a quadrotor at velocities of up to 21 m/s in the physical world, being on par in performance with the AC-MLP baseline.
}
In general, we show that our method can tackle challenging control tasks and achieves robust control performance for agile flight. 

However, there are some limitations to be mentioned and to be improved in the future.
\rev{
First, AC‑MPC relies on differentiable MPC~\cite{amos2018differentiable}, whose theoretical framework and gradient derivation cover input constrained problems only. 
Therefore, state constraints are not supported, since this solver doesn't implement them. Consequently, our controller inherits this limitation and can currently enforce input limits but not explicit state bounds.
This becomes particularly evident in safety-related tasks, such as collision avoidance or enforcing a speed limit. 
In such scenarios, simply penalizing violations as a term in the reward function does not guarantee that the constraints will be respected at deployment time. 
We believe that the work introduced in this paper highlights and motivates the practical value of developing a differentiable MPC solver that natively supports state constraints
Our results serve as strong encouragement for the numerical optimization community to further work in that direction.
In addition, AC-MPC may underperform or fail in certain practical scenarios. One such case is solver non-convergence. 
Since AC-MPC relies on solving an optimization problem at every control step, it is susceptible to failures if the problem becomes ill-conditioned or infeasible due to, e.g., extreme cost weights, poor initialization, or tight constraints. 
Such failures can result in invalid control actions and incorrect gradient updates during training. 
While this issue was rare in our experiments (especially after enforcing positive definiteness of the cost matrices) it remains a relevant concern.

Another key limitation is the scalability of the optimization block. The differentiable MPC~\cite{amos2018differentiable} solver used in this work is based on iterative LQR, which scales cubically with the state and input dimensions, and linearly with the horizon length.
As a result, control of high-dimensional, more complex systems or use of longer horizons may become impractical for real-time deployment. 
\revi{While all experiments in this work are conducted using a quadrotor platform with moderate dimensionality, training and deploying and extending our AC-MPC framework to more complex, high-dimensional robotic systems represents a promising direction for future research.
Such extensions would also motivate the development of more efficient differentiable solvers to further reduce training and inference times.}

Furthermore, AC-MPC assumes that the system dynamics can be expressed as an analytical formulation, and that they are differentiable, since gradients must be propagated through the MPC solver during training. 
This restricts the applicability of our method to systems with known, differentiable dynamics expressed in closed form.
In the absence of such models, the core assumptions needed to compute gradients through the solver no longer apply.

We believe that successfully tackling these challenges would unlock the demonstrated benefits of AC-MPC for a much wider range of real-world robotic applications, significantly broadening the impact of this architecture.
}

\rev{
A final point of discussion is the distinction between AC-MPC and standard MBRL methods.
First, standard MBRL is designed for scenarios where dynamics are unknown and must be learned from data, whereas the AC-MPC framework assumes access to a known, analytical dynamics model as a key component of its differentiable MPC module.
Second, most state-of-the-art MBRL methods are off-policy, leveraging a replay buffer for improved sample efficiency. 
A third key difference lies in reward design. 
AC-MPC benefits from the flexibility to learn from any high-level reward signal, irrespective of its shape or differentiability. 
In contrast, MBRL methods that rely on planning through the model, particularly those using gradient-based techniques, often require more carefully structured or differentiable reward functions to ensure stability and enable effective planning.

In our work we show that our approach enhances an on-policy algorithm (PPO) by integrating an optimization-based prior from MPC. 
Given this, our approach is best understood as a method to improve the properties of an on-policy algorithm, and the most sensible comparisons are against the components it seeks to integrate: the pure model-free approach (AC-MLP) and the classic optimization-based controller (MPC). 
This allows for a clear demonstration of the value added by our hybrid architecture.
}

\revi{While our results focus on agile quadrotor flight, we believe that the proposed method represents an important step in the direction of generalizability and out of distribution behaviour in RL.}
The proposed method demonstrates that modular solutions, which combine the advantages of learning-centric and model-based approaches, are becoming increasingly promising.
Our approach potentially paves the way for the development of more robust RL-based systems, contributing positively towards the broader goal of advancing AI for real-world robotics applications.

\balance
\IEEEtriggeratref{105}
\bibliographystyle{IEEEtran}
\bibliography{references}
\vspace*{-3.0\baselineskip}
\begin{IEEEbiography}[{\includegraphics[width=1in,height=1.25in,clip,keepaspectratio]{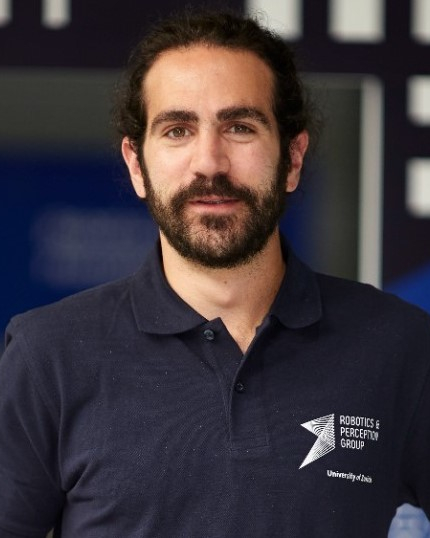}}]{Angel Romero} received a MSc degree in "Robotics, Systems and Control" from ETH Zurich in 2018.
Previously, he received a B.Sc. degree in Electronics Engineering from the University of Malaga in 2015.
He is currently working toward a Ph.D. degree in the Robotics and Perception Group at the University of Zurich, finding new limits in the intersection of machine learning, optimal control, and computer vision applied to super agile autonomous quadrotor flight under the supervision of Prof. Davide Scaramuzza.
\end{IEEEbiography}
\vspace*{-3.0\baselineskip}
\begin{IEEEbiography}[{\includegraphics[height=1.2in, keepaspectratio]{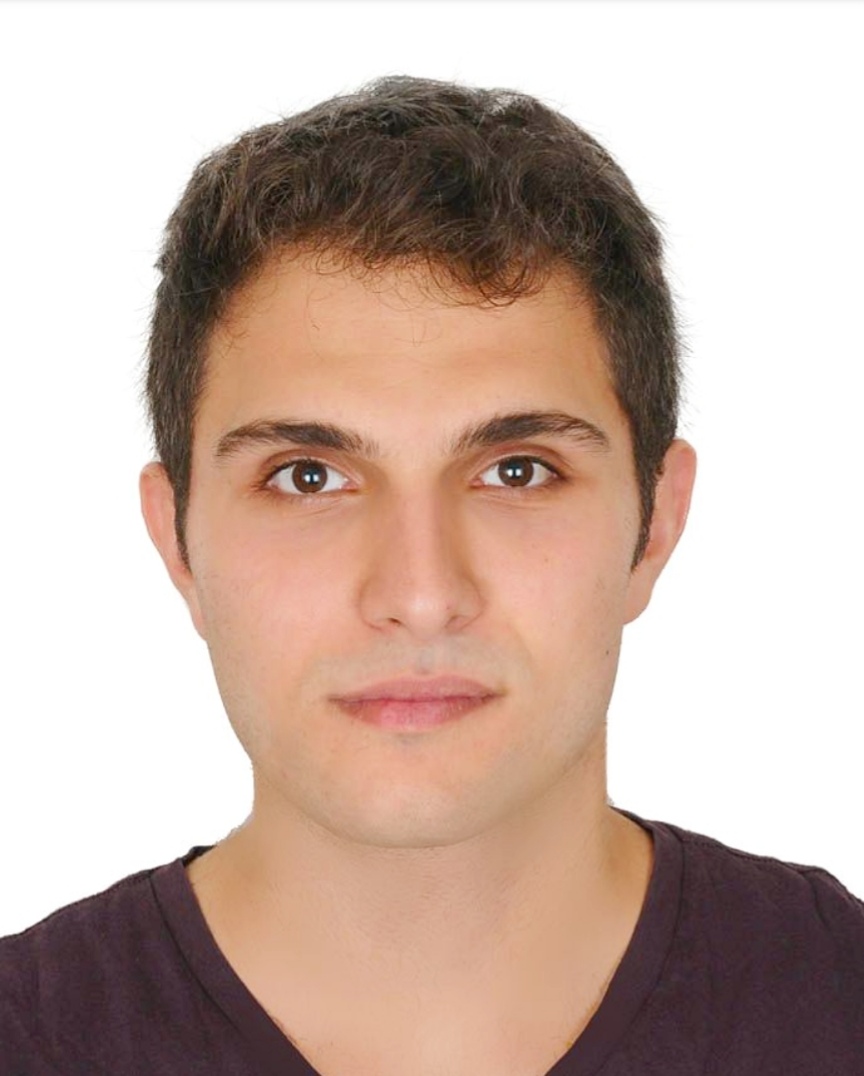}}]{Elie Aljalbout} received his Ph.D. degree from the Technical University of Munich, Germany, where he researched robot learning and manipulation at the Munich Institute of Robotics and Machine Intelligence. During his Ph.D., he also worked as a Research Scientist at the Volkswagen Group Machine Learning Research Lab until 2024. From 2024 to 2025, he was a Postdoctoral Researcher at the University of Zurich and an Associated Postdoctoral Researcher at the ETH AI Center. Since 2025, he has been an Embodied AI Researcher at Meta AI (FAIR), focusing on world models, latent action modeling, and foundation models for robotic manipulation and embodied agents. His research interests include large-scale pretraining for robotics, world models, vision-language-action models, reinforcement learning, optimal control, and self-supervised learning.
\end{IEEEbiography}
\vspace*{-2.0\baselineskip}
\begin{IEEEbiography}[{\includegraphics[height=1.2in, keepaspectratio]{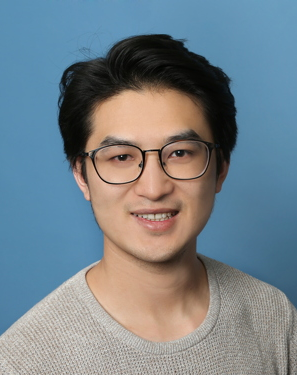}}]{Yunlong Song} obtained the M.Sc. degree in Information and Communication Engineering from Technical University of Darmstadt in 2018. In 2024, he completed his Ph.D. in the Robotics and Perception Group at the University of Zurich under the supervision of Prof. Davide Scaramuzza. His research interests include reinforcement learning, machine learning, and robotics. 
\end{IEEEbiography}
\vspace*{-2.5\baselineskip} %
\begin{IEEEbiography}[{\includegraphics[width=1in,height=1.25in,clip,keepaspectratio]{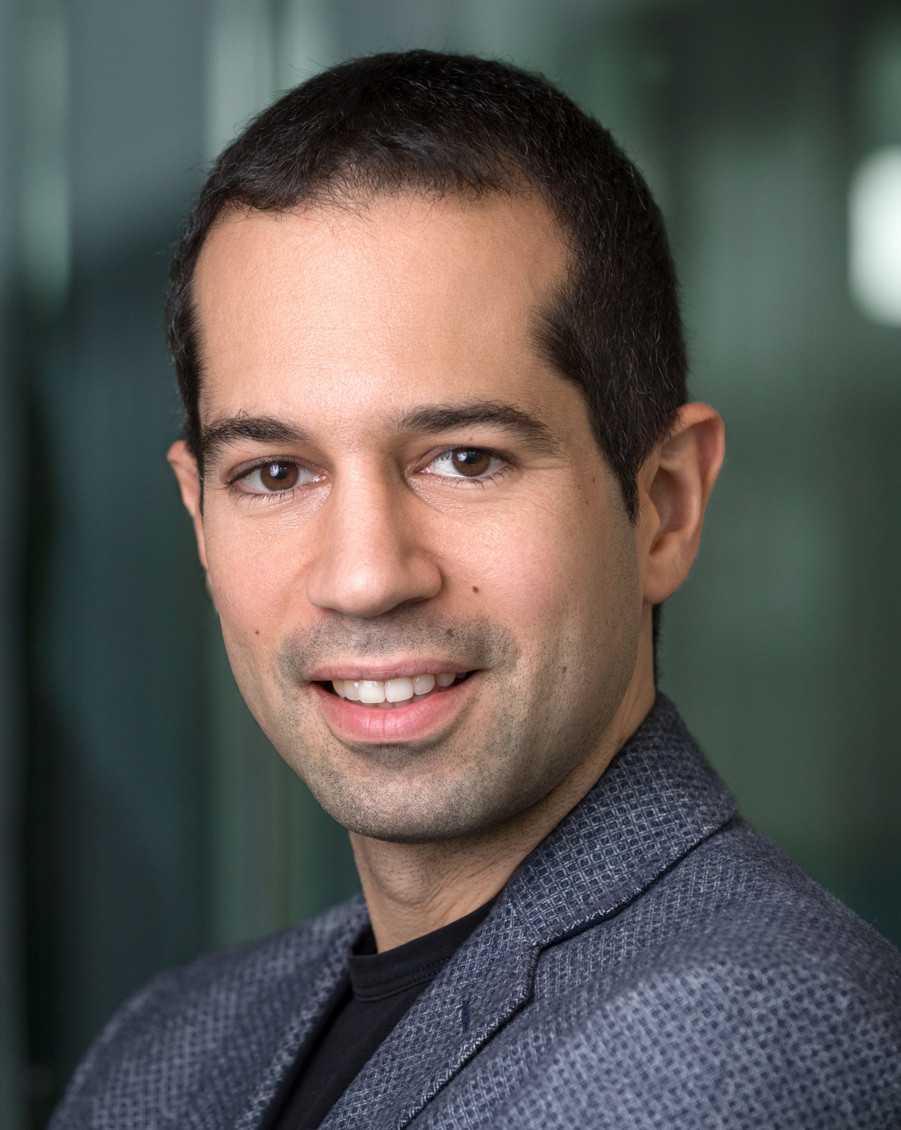}}]{Davide Scaramuzza} is a Professor of Robotics and Perception at the University of Zurich. He did his Ph.D. at ETH Zurich, a postdoc at the University of Pennsylvania, and was a visiting professor at Stanford University. His research focuses on autonomous, agile microdrone navigation using standard and event-based cameras. He pioneered autonomous, vision-based navigation of drones, which inspired the navigation algorithm of the NASA Mars helicopter and many drone companies. He contributed significantly to visual-inertial state estimation, vision-based agile navigation of microdrones, and low-latency, robust perception with event cameras, which were transferred to many products, from drones to automobiles, cameras, AR/VR headsets, and mobile devices. In 2022, his team demonstrated that an AI-controlled, vision-based drone could outperform the world champions of drone racing, a result that was published in Nature. He is a consultant for the United Nations on disaster response, AI for good, and disarmament. He has won many awards, including an IEEE Technical Field Award, the IEEE Robotics and Automation Society Early Career Award, a European Research Council Consolidator Grant, a Google Research Award, two NASA TechBrief Awards, and many paper awards. In 2015, he co-founded Zurich-Eye, today Meta Zurich, which developed the world-leading virtual-reality headset Meta Quest. In 2020, he co-founded SUIND, which builds autonomous drones for precision agriculture. Many aspects of his research have been featured in the media, such as The New York Times, The Economist, and Forbes.
\end{IEEEbiography}

\end{document}